\documentclass{article}

\usepackage{microtype}
\usepackage{graphicx}
\usepackage{subfigure}
\usepackage{booktabs} 
\usepackage{amsfonts}
\usepackage{bm}
\usepackage{color}
\usepackage{makecell}

\usepackage{balance}

\usepackage{hyperref}

\usepackage[accepted]{mlsys2020}

\mlsystitlerunning{Submission and Formatting Instructions for MLSys 2020}

\begin{document}

\newcommand{\fetch}{{\tt fetch}}
\newcommand{\update}{{\tt update}}

\newcommand{\Comment}[1]{{\color{red} comment:}\emph{#1}}

\newcommand{\nrnd}[1]{{\color{blue} #1}}
\newcommand{\srnd}[1]{{\color{red} #1}}

\twocolumn[
\mlsystitle{
Mixed-Precision Embedding using a Cache}



\mlsyssetsymbol{equal}{*}

\begin{mlsysauthorlist}
\mlsysauthor{Jie (Amy) Yang}{equal,fb}
\mlsysauthor{Jianyu Huang}{equal,fb}
\mlsysauthor{Jongsoo Park}{fb}
\mlsysauthor{Ping Tak Peter Tang}{fb}
\mlsysauthor{Andrew Tulloch}{fb}
\end{mlsysauthorlist}

\mlsysaffiliation{fb}{Facebook Inc., Menlo Park, California, USA}

\mlsyscorrespondingauthor{Jie (Amy) Yang}{amyyang@fb.com}

\mlsyskeywords{Machine Learning, MLSys}

\vskip 0.4in

\begin{abstract}
In recommendation systems, practitioners observed that increase in the number of embedding tables and their sizes often leads to significant improvement in model performances. Given this and the business importance of these models to major internet companies, embedding tables for personalization tasks have grown to terabyte scale and continue to grow at a significant rate. Meanwhile, these large-scale models are often trained with GPUs where high-performance memory is a scarce resource, thus motivating numerous work on embedding table compression during training.
We propose a novel change to embedding tables using a cache memory architecture, where the majority of rows in an embedding is trained in low precision, and the most frequently or recently accessed rows cached and trained in full precision. The proposed architectural change works in conjunction with standard precision reduction and computer arithmetic techniques such as quantization and stochastic rounding.
For an open source deep learning recommendation model (DLRM) running with Criteo-Kaggle dataset, we achieve 3$\times$ memory reduction with INT8 precision embedding tables and full-precision cache whose size are 5\% of the embedding tables, while maintaining accuracy.
For an industrial scale model and dataset, we achieve even higher $>$7$\times$ memory reduction with INT4 precision and cache size 1\% of embedding tables, while maintaining accuracy, and 16\% end-to-end training speedup by reducing GPU-to-host data transfers.
\end{abstract}
]

\printAffiliationsAndNotice{\mlsysEqualContribution} 

\section{Introduction}\label{sec:introduction}
Machine learning and deep learning in particular has been tremendously successful in tackling tasks that were traditionally considered to require human intelligence or intuition. Deep learning typically accomplishes this feat with massive computations in the Euclidean space on matrices of real numbers. While inputs to many deep learning tasks are naturally represented as matrices of continuous numerical values such as image classification and detection, in the cases where input data is discrete in nature such as  categorical features and vocabularies, embedding is at present the de facto technique that maps a discrete set into the continuous Euclidean space. In essence, a good embedding maps discrete entries into $\mathbb{R}^d$ in such a way that preserves natural relationship via the basic operations in Euclidean space. 





Naturally, embedding table is an important component of recommendation systems that are important to internet companies \cite{cheng16_deepwide,wang17_crossnet,hazelwood2018,DLRM19}, which rely heavily on categorical features to deliver personalized contents based on user-item interactions. Such models customarily contain hundreds of embedding tables representing different users, characteristics of commercial items, news articles' topics, and are commonly tens of gigabytes in size in commercial ML models \cite{hazelwood2018,Park2018}.

Moreover, industry generally believes that further growing the embedding table size will improve the models' predictive performance. However, such scaling of embeddings poses a challenge in scaling the training system's memory and computation capacity accordingly. In GPU training systems with highly parallel execution but limited high-bandwidth memory capacity, the cost of training can be significantly increased as embeddings trend towards billions scale. Due to industry's interest in recommendation models, accommodating the intense memory requirement of embeddings is essential in scaling high-performance training systems.

\begin{figure}
    \centering
    \includegraphics[width=3.2in]{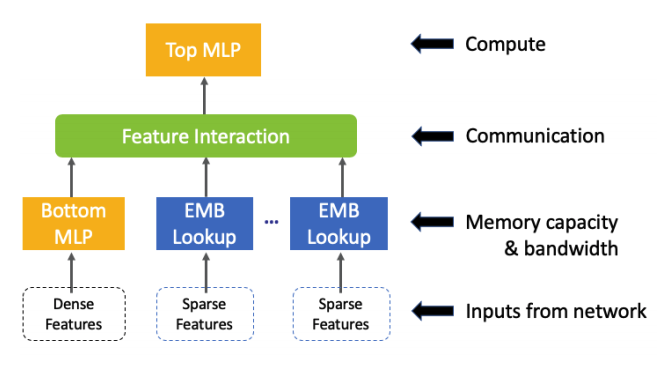}
    \caption{Deep Learning Recommendation Model (DLRM) \cite{DLRM19}}
    \label{fig:DLRM}
\end{figure}

Figure~\ref{fig:DLRM} captures the structure of a representative Deep Learning Recommendation Model (DLRM)~\cite{DLRM19} developed for personalization tasks. Dense and sparse features are computed via multi-layer perceptron (MLP) and embedding lookups, then joined in sparse-dense interaction and top MLP to compute the final click-through rate prediction. Large embeddings are partitioned across multiple GPUs following model-parallelism with no replication across device. This paper focuses on training large embeddings with reduced memory footprint while maintaining accuracy. Training acceleration via dense computation optimization is not the main subject of this study.

In this work, we propose to train embeddings in low precision with the addition of a small full-precision cache for the most recently or frequently accessed rows. Depending on the specific cache replacement policy in play, some embedding rows may behave as if they were trained in full precision most or even all of the time. Previous work~\cite{zhang2018} shows that we can train embeddings in FP16 with stochastic rounding while maintaining neutral accuracy. We show that with a small high-precision cache, we can train embeddings with INT8 and INT4 precision, further narrowing the bitwidths and thus the memory consumption at little to no cost of accuracy. While caching is commonly adopted in GPU training systems to utilize CPU host memory~\cite{gpu_hierarchical}, we reuse the same architecture with mixed-precision to achieve the purpose of reduced memory footprint for embeddings and maintain a neutral training accuracy. 

Specifically, this paper makes the following contributions:

\begin{itemize}
\item We propose mixed-precision embedding training with a full-precision cache, and show its effectiveness and generality in reducing embeddings memory (3$\times$ memory reduction on an open source DLRM model and 7$\times$ memory reduction on an industrial scale model) usage during training and maintaining model accuracy.
\item We compare different cache replacement policies: least recently used (LRU), least frequently used (LFU), and the effect of cache associativity on model accuracy and cache hit rate. 
\item We explore the effect of varying cache sizes on model accuracy with embeddings trained in low precision. While model accuracy improves with large cache sizes, there is diminishing return in accuracy recovery with increasingly large cache size.
\item We study the effect of rounding modes on model accuracy with and without cache, and show that stochastic rounding consistently outperforms round-to-nearest in low-precision embedding training by getting rid of systematic rounding bias.
\item We achieve 16\% end-to-end training speed improvement on an industrial scale recommendation system using our optimized  full-precision cache implementation. With 95\% cache hit rate, the embedding lookup throughput is improved by 25.8$\times$ over that of unified memory.
\end{itemize}


The rest of paper is organized as follows. Section \ref{sec:related_works} reviews prior work on embedding table compression techniques. Section \ref{sec:cache} explains our approach and different cache replacement policies. Section \ref{sec:cpu} discusses the experiments and results with CPU emulation and open source DLRM model. The results support our focus on an associative cache GPU implementation described in Section \ref{sec:gpu}. The GPU implementation enables us to experiment with large-scale data set that would otherwise be prohibitive on CPU. Section \ref{sec:conclusion} recaps our result and discusses future work.


\section{Related Works}\label{sec:related_works}

A sample of works showing embedding in action are \cite{mikolov2013distributed,pennington2014glove,liu2015topical,peters2018dissecting,liu2019roberta} for natural language processing, \cite{vasileva2018learning, barz2019hierarchy} for computer vision, and other areas such as knowledge representation and biology \cite{wu2018fonduer, chiu2016train}.

We follow the conceptual classification termed in \cite{ginart2019} of algorithmic and architectural compression of embedding tables. Some compression algorithms apply post processing methods on the embedding tables such as low-rank SVD and other factorizations \cite{Bhavana2019,Andrews2015}, sparsification \cite{Sattigeri,Sun2016} and quantization \cite{guan2019}. Other compression algorithms aim at arriving at easily compressible tables through training. These include quantization or compression-aware training \cite{alvarez2017compression,Park2018,Naumov2018reg,Elthakeb2019} and gradual pruning \cite{frankle2018lottery}. These techniques reduce   the memory requirement at inference time but use the uncompressed embedding tables during training.

Alternatively, compressed architectures use alternative representation in the ML model architecture of the embedding tables other than the standard 2D-array of FP32 floating-point numbers. Hence memory saving is realized even during training. Along this line, representing the tables in low-precision data type is a natural strategy. Works along this line mainly need to maintain the models' accuracy despite the models' parameters have reduced precision. Notable works using this approach include \cite{gupta2015limited,chen2017fxpnet,de2018high,wu2018integers,zhang2018,kalamkar2019}. We note that some of the just referenced works try to reduce the memory footprint of weight parameters so as to improve compute performance and not that of embedding table size. Some other works explore more structural changes to the embedding table such as tensor train representation \cite{Khrulkov2019}, mixed-dimension factorization \cite{ginart2019} or vector quantization \cite{fan2020}.

Instead of using algebraic factorization, we use memory hierarchy and varying precision to enable compression. Like the other compression architecture work, the compression benefits training as well as inference. The sequel of this paper explains our design and reports the results on extensive experiments with our CPU and GPU implementations running different models on multiple data sets with different cache size/precision/policy combinations.

\section{Mixed-Precision Embedding Table with a Cache}\label{sec:cache}

A common embedding table $T$ consists of $N$ rows of $d$-dimensional vectors of FP32 numbers. Each row corresponds to some categorical features, for example a particular user, and any production grade ML model such as a recommendation system can consist of hundreds of such tables. 
Expanding the capacity of embedding tables -- by increasing the total number of tables or dimensions of individual ones -- can increase model performance, motivating active research in embedding table compression that does not sacrifice the tables' representation capacities. 

To explain our compression method of mixed-precision with a cache, it suffices to consider the workflow involving a single row of one table $T$ during a training iteration. During a forward pass, a particular index-$i$ row $\mathbf{x}$ of $T$ is needed and obtained by a fetch operator $\mathbf{x} \gets \fetch(T,i)$. This row will participate in the model evaluation as input to some layers and eventually contribute to the loss function $L$. During the corresponding backward pass, the gradient of $L$ with respect to $\mathbf{x}$, $\frac{\partial L}{\partial \mathbf{x}}$ is computed and $\mathbf{x}$ updated via $\mathbf{x} \gets \mathbf{x} - \eta \frac{\partial L}{\partial \mathbf{x}}$ for example with a simple SGD algorithm with learning rate $\eta$. This update to the embedding table is denoted as $\update(T,i,\mathbf{x})$. 

Our method has two main components. The first is the use of a reduced-precision embedding table similar to \cite{zhang2018}. Table $T$ is in a precision lower than FP32. During a forward pass, the $\fetch$ operator $\mathbf{x} \gets \fetch(T,i)$ up-converts the row in its low-precision representation to FP32. Computations are carried out in FP32 including the computation $\mathbf{x} \gets \mathbf{x}-\eta\frac{\partial L}{\partial \mathbf{x}}$. The update step $\update(T,i,\mathbf{x})$ uses either a round-to-nearest or stochastic rounding \cite{zhang2018} approach to down convert $\mathbf{x}$ into the precision of $T$'s entries. \cite{zhang2018} applies this approach to an FP16 embedding table $T$ and uses stochastic rounding to store the updated row back into $T$. 


Our second component augments the low-precision embedding table $T$ with a cache $C$ of FP32 entries. $C$ only has $n$ rows, which is a fraction of $T$'s row dimension $N$. Similar to cache memory architecture, an embedding row may be residing both in the cache $C$ and the table $T$, albeit with different precision. We use and preserve the high precision version whenever possible. 

Combining these two components, the method is encapsulated by the $\fetch$ and $\update$ operators as $\mathbf{x} \gets \fetch(T,C,i)$ and $\update(T,C,i,\mathbf{x})$. The operator $\fetch$ returns the higher-precision version if the row $i$ resides in cache, or return the up converted result of the lower-precision version in $T$. In updating the table and cache with the updated row $\mathbf{x}$, the $\update$ operator checks that if this row is cache resident, simply replaces the row in cache with the updated $\mathbf{x}$. Otherwise, one of two scenarios happen that is determined by the cache policy in effect. Either $\mathbf{x}$ does not have priority to evict the current cache content whose space it conflicts with, or that it does. In the former, $\mathbf{x}$ is placed in the cache verbatim while the evicted row is placed in the table $T$ after being down converted. In the latter, one simply down converts $\mathbf{x}$ and places it in $T$. 

Our design allows four knobs as follows.
\begin{itemize}
    \item {\bf Precision:} The embedding table $T$ contains entries in precision lower than FP32. We allow the IEEE 16-bit floating-point values FP16 as well as quantization into INT8, INT4, as well as INT2 precision. For integer quantization, each row of the embedding values share one pair of quantization parameters $(s,b)$, scale and bias, represented in FP32.
    We use row-wise uniform min-max quantization in mapping real values to unsigned quantization domain:
    \[ x_{\rm quantize} = \frac{x-b}{s} \]
    where b=min({\textbf x}) and s=(max({\textbf x})-min({\textbf x}))/($2^N-1$). The quantization parameters are chosen so that the quantized embedding values will be in the corresponding range of the unsigned integer datatype: $[0,3]$, $[0,15]$ and $[0,255]$ for INT2, INT4 and INT8 quantization, respectively.
    
    \item {\bf Rounding:} When a FP32 value has to be placed into a low-precision embedding table, a precision down conversion occurs. Let $x$ be the FP32 value in question. There is a pair of neighboring numbers $(x_-,x_+)$ in the quantized low-precision domain such that $x_- \le x \le x_+$ (that is, they differ by one ``unit in last place'' in FP16, or are consecutive integers in the case of quantization). We allow two rounding options: round to nearest and stochastic rounding. If \emph{round to nearest} is used, then the number in $\{x_-,x_+\}$ that is closest to $x$ is returned. In case of a tie, the one with even parity (least significant bit equals 0) is chosen. If \emph{stochastic rounding} is used, then one of $x_-$ and $x_+$ is chosen randomly with a Bernoulli distribution so that the expected value (average) is in fact $x$. 
    
    \item {\bf Cache Structure:} Let an uniform-dimension embedding table $T$ has $N$ rows with indices from $0$ to $N-1$, and each row has dimension [1, d]. The cache $C$ has $n$ sets each of which can hold $\alpha$ FP32 embedding rows. The cache size is thus $4\alpha n d$ bytes. We restrict ourselves to $\alpha \ge 1$ being an integral power of 2 and cache sizes chosen such that $\alpha, n$ are integers. A hash function
    \[ h:\{0,1,\ldots,N-1\} \rightarrow \{0,1,\ldots,n-1\}
    \]
    is chosen and row $i$ of $T$ is mapped to the set $h(i)$ of the cache. When $\alpha=1$, we have a direct mapped cache; otherwise, we have a set associative cache. 
    \item {\bf Cache Replacement Policy:} When we attempt to place a non-cache resident row of index $i$ into the cache and the set $h(i)$ is already fully occupied, the replacement policy dictates the appropriate action. We maintain a priority value of each row which can be either access frequency or timestamp of last access, depending on the replacement policy. We also keep track of the lowest priority value of the current cache residents for each set $s$, $0 \le s < n$. When the FP32 Row-$i$ is to be stored, our policy puts the row in the full-precision cache if and only if its priority value is higher than the lowest priority value of the residents. In this case, the lowest priority row in the cache will be evicted -- down converted by a rounding method of choice and placed back into the table $T$. Otherwise, Row-$i$ bypasses the cache, and after it is updated in FP32, is put into $T$ with down conversion. We have two specific policies specified by their respective priority value calculations. 
    \begin{enumerate}
        \item Least Frequently Used (LFU). Each embedding row carries with it an access count as the priority value. Access count is a popularity measure; LFU policy naturally let the more popular rows maintain higher precision based on the assumption that the more frequently used rows pull heavier weight on model accuracy. One drawback of this policy is that access count requires extra memory per row, whose size grows linearly with the number of rows in embeddings.
        
        \item Least Recently Used (LRU). Each cached row carries the last access timestamp as the priority value. This policy lets those rows that are often accessed within a time window to maintain higher precision in that duration. 
        Another advantage over LFU is that this priority value needs not be explicitly maintained at all in the case of a direct mapped cache, as the cache resident entry is always evicted in case of a conflict.
    \end{enumerate}
    
\end{itemize}
We will report on our experiments and results in the next sections.

\section{CPU Experiments with DLRM and Kaggle Dataset}\label{sec:cpu}

We use a CPU emulation to explore our algorithm design space with an open source model and dataset. Section \ref{sec:gpu} will discuss an implementation in a GPU training system and performance evaluation with an industrial-scale internal model and dataset. The main emulation aspect here is that we do not keep two physical spaces for the embedding table and the cache, but rather use a single FP32 array to emulate a single low-precision table with FP32 cache. We use the simple technique of fake quantization, detailed below, to maintain faithful behavior of low-precision numerics and experimented with all combinations of the four precisions and two rounding methods (discussed in section 3) on each of the following cache settings: direct-mapped cache with LFU and LRU policies and set associative LFU cache with associativity $\alpha = 2^k$, $1\le k \le 5$.


\subsection{Low-Precision Emulation}
We implemented a custom sparse AdaGrad optimizer that uses one high-precision tensor to store the embedding weights, and keep a list of row indices of the current cache residents. All embedding gradient updates are applied in full precision. When Row-$i$ in table $T$ is supposed to be stored in low precision, we apply fake quantization (quantization followed by dequantization) to row $T[i,:]$ to provide a numerically faithful emulation: In the case of FP16, fake quantization of a FP32 value $x$ is simply the operation
\[
{\tt FP16\_to\_FP32}\left({\tt FP32\_to\_FP16} (x)\right).
\]
With the above conversion, while the result physically remains in FP32, its precision is that of FP16. Fake integer quantization of an FP32 value $x$ similarly returns an FP32 object but whose precision is the same as if integer quantization was performed. Algorithm \ref{alg:cpu-rw-int-quant} shows the fake quantization details.



\begin{algorithm}[tb]
   \caption{Fake Row-wise Integer ${\tt IntN}$ Quantization\\
   Parameters: Uniform Min-Max}
   \label{alg:cpu-rw-int-quant}
\begin{algorithmic}
   \STATE {\bfseries Input:} embedding row $\mathbf{r}$, embedding dimension $d$, bitwidth $N$
   \STATE Compute $b \gets \min(\mathbf{r})$, $s \gets (\max(\mathbf{r})-b)/(2^N-1)$
   \FOR{$i=0,1,..., d-1$}
   \STATE $q \gets {\rm round\_to\_int}((\mathbf{r}[i]-b) / s)$ (nearest or stochastic)
   \STATE $\mathbf{r}[i] \gets q \times s + b$
   \ENDFOR
\end{algorithmic}
\end{algorithm}

\subsection{Cache Implementations}
Numerical faithfulness is paramount in our study; thus we show the exact numerical steps on our single-array emulation of an embedding table and its cache. Here we have a single table $T$ where some rows correspond to cache residents in FP32 precision and the others non-cache residents in the corresponding low precision.

At any given iteration, let $I_U$ be the set of embedding row indices to be updated by a scaled gradient, $U$. Let $PV[k]$ be the priority value for the embedding row $k$, either for LRU or LFU. Note that in the case of $\alpha=1$ (direct-mapped cache) for LRU, $PV$ is not required as all accessed rows will be the most recent by default and no priority comparison is necessary. Tag indices $I_C$ are maintained to keep the row ids of cache residents for each of the $\alpha$ rows residing in the set $h(i)$, with $h$ being the hash function. Algorithm \ref{alg:cpu-assoc} gives the numerical details of our cache implementation.

\begin{algorithm}
   \caption{Update $\alpha$-Associative Cache}
   \label{alg:cpu-assoc}
\begin{algorithmic}
   \STATE {\bfseries Input:} embedding $T$, tag indices $I_C$ of current cached rows, indices to update $I_U$, hash function $h$, gradient update $U$, priority value $PV$ (non existent for $\alpha=1$, direct-mapped LRU)
   \FOR{$i$ in $I_U$}
   \STATE Update $PV[i]$ according to LRU or LFU
   \STATE $s\gets h(i)$
   \STATE $T[i,:] \gets T[i,:] + U[i,:]$ \# apply gradient updates
   \IF{$i \notin I_C[s]$}
     \STATE $j \gets {\rm arg min}\{PV[k] | k \in I_C[s]\}$
     \IF{$PV[i] \le PV[j]$ (False if $\alpha=1$ and LRU)}
     \STATE \# bypass
     \STATE $T[i,:] \gets {\tt fake\_quantize}(T[i,:])$
     \ELSE 
     \STATE \# replace
     \STATE $T[j,:] \gets {\tt fake\_quantize}(T[j,:])$
     \STATE replace $j$ by $i$ in the set $I_C[s]$
     \ENDIF
   \ENDIF
   \ENDFOR
\end{algorithmic}
\end{algorithm}

\subsection{Experiments and Results}
We use Deep Learning Recommendation Model (DLRM)~\cite{DLRM19} with Criteo-Kaggle 7D Ads Display Challenge dataset\footnote{http://labs.criteo.com/2014/02/kaggle-display-advertising-challenge-dataset/}~\cite{kaggle-7d} for model accuracy evaluations with different low-precision and cache implementations. The accuracy metric is based on the relative change to the test accuracy of the resulting low-precision models using the full FP32 model as reference. The precise definition of our metric is
\begin{equation}
\label{eq:accuracy_drop}
    \hbox{Accuracy Drop \%} = \frac{{\rm Acc}_{\rm FP32} - {\rm Acc}_{\rm lowprec}}{{\rm Acc}_{\rm FP32}} \cdot 100
\end{equation}
In particular, if the low-precision model in fact has higher test accuracy, our metric becomes negative. Our goal is to keep this number below roughly $0.02\%$.

Our benchmark model has 26 embedding tables with the following sizes shown in Table \ref{cpu:dlrm-emb-sizes}. We configure all 26 embedding tables to use row dimension 128 (total 4.3B parameters), and only apply low-precision with or without caching technique to tables with more than 1K rows for all experiments. Embedding tables with less than 1K rows are trained in FP32. 

\begin{table}[h]
\caption{DLRM benchmark model embedding sizes}
\label{cpu:dlrm-emb-sizes}
\vskip 0.15in
\begin{center}
\begin{small}
\begin{sc}
\begin{tabular}{rrrr}
\toprule
4 & 4 & 11 & 16  \\
18 & 24 & 28 & 105  \\
306 & 584 & 634 & 1,461  \\
2,173 & 3,195 & 5,653 & 5,684  \\
12,518 & 14,993 & 93,146 & 142,572  \\
286,181 & 2,202,608 & 5,461,306 & 7,046,547  \\
8,351,593 & 10,131,227 \\
\bottomrule
\end{tabular}
\end{sc}
\end{small}
\end{center}
\vskip -0.1in
\end{table}

\subsubsection{Accuracy Recovery with High-precision Cache}
\label{subsec:results_cache}
We first evaluate accuracy loss of training with low-precision embeddings without a high-precision cache. Table \ref{tab:cpu-no-cache-drop} shows the relative test accuracy drop of training embeddings in various low precisions with nearest and stochastic rounding. We observe that neutral accuracy can be achieved with FP16 with stochastic rounding without high-precision caching; but accuracy decreases significantly with progressively narrower bitwidth. The accuracy boost from stochastic rounding is consistent with previous work on low-precision embedding training~\cite{zhang2018}, but not nearly enough for embedding tables represented by 8 or fewer bits. This loss of accuracy was part of the motivation for our high-precision cache.

\begin{table}[h]
\caption{Test Accuracy Drop without Cache in \%}
\label{tab:cpu-no-cache-drop}
\begin{center}
\begin{small}
\begin{sc}
\begin{tabular}{lcccr}
\toprule
~   &   fp16 & int8 &  int4   &  int2    \\ 
\midrule
nearest  &  0.047   &   0.549 & 1.080  &  1.454 \\ 
stochastic  & {\bf -0.010} & 0.077  &  0.591  & 1.037 \\
\bottomrule
\end{tabular}
\end{sc}
\end{small}
\end{center}
\vskip -0.1in
\end{table}


We experiment with direct-mapped high-precision LFU, LRU cache, and set associative LFU cache with varying sizes: 5\%, 10\%, 30\%, and 50\% (of the original FP32 table) on top of low-precision tables with nearest/stochastic rounding. 32-Way LFU gives the best accuracy results and as shown in Table \ref{tab:cpu_assoc_lfu}: we achieve neutral accuracy with INT8, INT4, and INT2 embeddings (with stochastic rounding) at cache sizes of 5\%, 30\%, and 50\%, respectively.

\begin{table}[h]
\caption{High-precision 32-Way LFU cache with varying cache sizes recover accuracy of low-precision embedding tables}
\label{tab:cpu_assoc_lfu}
\begin{center}
\begin{scriptsize}    
\begin{sc}
\begin{tabular}{l | c c c }
\multicolumn{1}{c|}{
\makecell{Cache Size\\ 
\% of}} &
\multicolumn{3}{c}{
\makecell{Test Accuracy Drop in \% \\
Rounding: \nrnd{Nearest}/\srnd{Stochastic}}
} \\
\multicolumn{1}{c|}{Table} & Int8 & Int4 & Int2 \\ \hline
5\% & \nrnd{0.074}/\srnd{\bf -0.014} & \nrnd{0.251}/\srnd{0.110} & \nrnd{0.376}/\srnd{0.275} \\
10\% & \nrnd{0.033}/\srnd{\bf -0.013} & \nrnd{0.157}/\srnd{0.073} & \nrnd{0.257}/\srnd{0.196} \\
30\% & \nrnd{\bf 0.000}/\srnd{\bf -0.021} & \nrnd{0.043}/\srnd{\bf -0.009} & \nrnd{0.086}/\srnd{0.077}\\
50\% & \nrnd{\bf -0.005}/\srnd{\bf -0.008} & \nrnd{\bf 0.018}/\srnd{\bf -0.004} &  \nrnd{0.042}/\srnd{\bf 0.025} \\

\end{tabular}
\end{sc}
\end{scriptsize}
\end{center}
\end{table}

Tables \ref{tab:cpu_assoc_lfu}, \ref{tab:cpu_direct_lru}, and \ref{tab:cpu_direct_lfu} show that the presence of a high-precision cache with as low as 5\% capacity of the original table size, regardless of cache replacement policy, recovers accuracy significantly for various lower precision levels with either nearest or stochastic rounding. With 5\% high-precision cache, we recover 70-80\% of accuracy drop from low-precision embeddings for 8 bits and lower.


\begin{table}[h]
\caption{High-precision LRU cache with varying cache sizes recover accuracy of low-precision embedding tables}
\label{tab:cpu_direct_lru}
\begin{center}
\begin{scriptsize}    
\begin{sc}
\begin{tabular}{l | c c c c }
\multicolumn{1}{c|}{
\makecell{Cache\\ Size\\ 
\% of}} &
\multicolumn{4}{c}{
\makecell{Test Accuracy Drop in \% \\ Cache: LRU, Direct Mapped\\
Rounding: \nrnd{Nearest}/\srnd{Stochastic}}
} \\
\multicolumn{1}{c|}{Table} & FP16 & Int8 & Int4 & Int2 \\ \hline

5\% & \nrnd{\bf 0.005}/\srnd{\bf -0.005} & \nrnd{0.180}/\srnd{\bf 0.010} & \nrnd{0.520}/\srnd{0.259} & \nrnd{0.602}/\srnd{0.872} \\
10\% & \nrnd{\bf 0.003}/\srnd{\bf 0.004} & \nrnd{0.144}/\srnd{\bf 0.008} & \nrnd{0.430}/\srnd{0.212} & \nrnd{0.761}/\srnd{0.536} \\
30\% & \nrnd{\bf 0.004}/\srnd{\bf -0.008} & \nrnd{0.093}/\srnd{\bf 0.006} & \nrnd{0.321}/\srnd{0.164} & \nrnd{0.582}/\srnd{0.425} \\
50\% & \nrnd{\bf 0.006}/\srnd{\bf -0.016} & \nrnd{0.043}/\srnd{\bf -0.004} & \nrnd{0.246}/\srnd{0.143} & \nrnd{0.462}/\srnd{0.352} \\

\end{tabular}
\end{sc}
\end{scriptsize}
\end{center}
\end{table}


\begin{table}[h]
\caption{High-precision LFU cache with varying cache sizes recover accuracy of low-precision embedding tables}
\label{tab:cpu_direct_lfu}
\begin{center}
\begin{scriptsize}    
\begin{sc}
\begin{tabular}{l | c c c c }
\multicolumn{1}{c|}{
\makecell{Cache\\ Size\\ 
\% of}} &
\multicolumn{4}{c}{
\makecell{Test Accuracy Drop in \% \\ Cache: LFU, Direct Mapped\\
Rounding: \nrnd{Nearest}/\srnd{Stochastic}}
} \\
\multicolumn{1}{c|}{Table} & FP16 & Int8 & Int4 & Int2 \\ \hline

 5\% & \nrnd{\bf -0.006}/\srnd{\bf -0.001} & \nrnd{0.087}/\srnd{\bf 0.014} & \nrnd{0.333}/\srnd{0.147} & \nrnd{0.533}/\srnd{0.423} \\
 10\% & \nrnd{\bf -0.011}/\srnd{\bf -0.003} & \nrnd{0.063}/\srnd{\bf 0.008} & \nrnd{0.270}/\srnd{0.138} &\nrnd{0.430}/\srnd{0.375} \\
 30\% & \nrnd{\bf -0.004}/\srnd{\bf -0.003} & \nrnd{0.044}/\srnd{\bf 0.009} & \nrnd{0.175}/\srnd{0.090} & \nrnd{0.280}/\srnd{0.243} \\
 50\% & \nrnd{\bf -0.018}/\srnd{\bf -0.008} & \nrnd{\bf 0.014}/\srnd{\bf 0.008} & \nrnd{0.136}/\srnd{0.056} & \nrnd{0.217}/\srnd{0.193} \\

\end{tabular}
\end{sc}
\end{scriptsize}
\end{center}
\end{table}

\begin{figure}[h!]
\begin{tabular}{cc}
  \includegraphics[width=40mm]{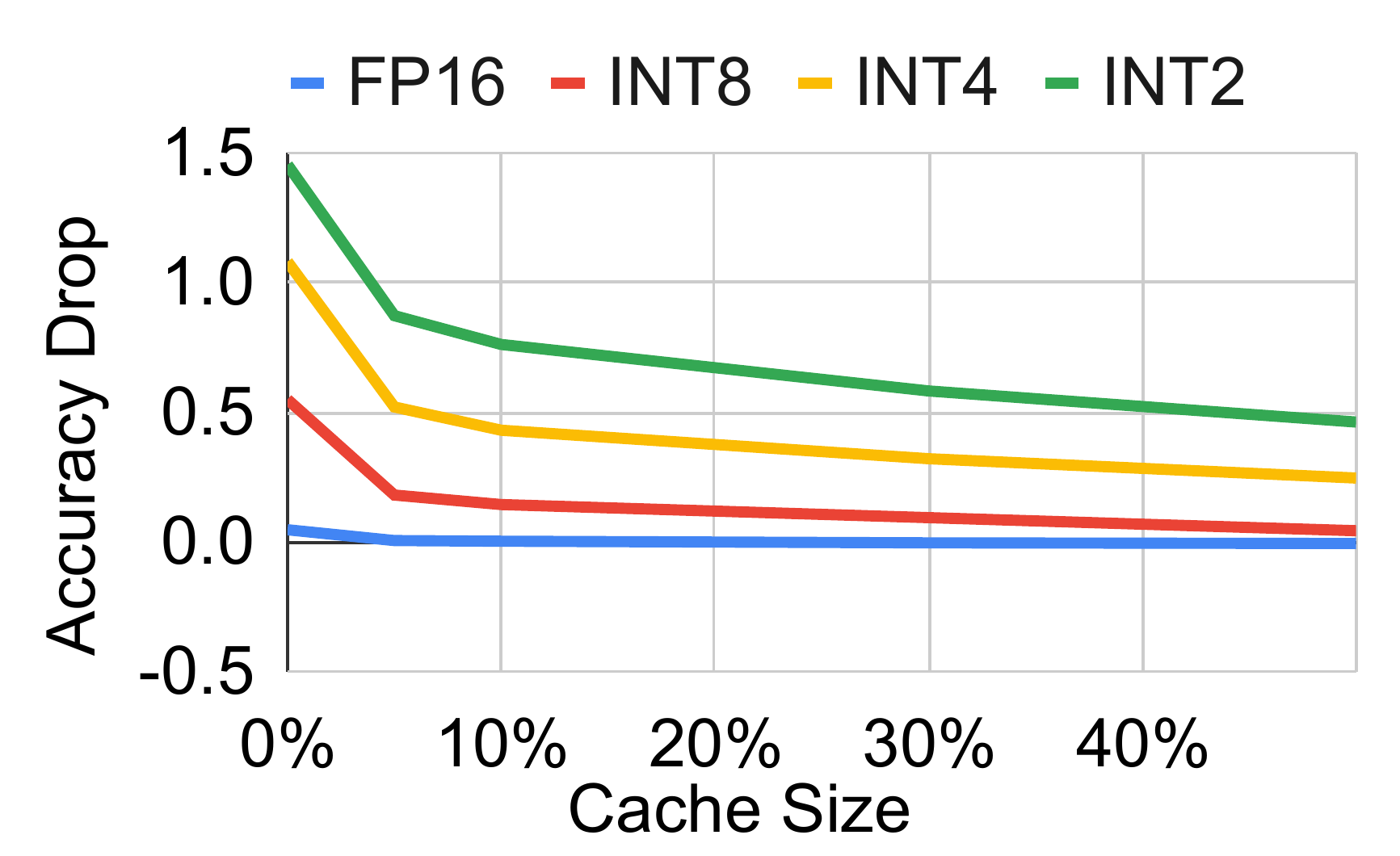} &   \includegraphics[width=40mm]{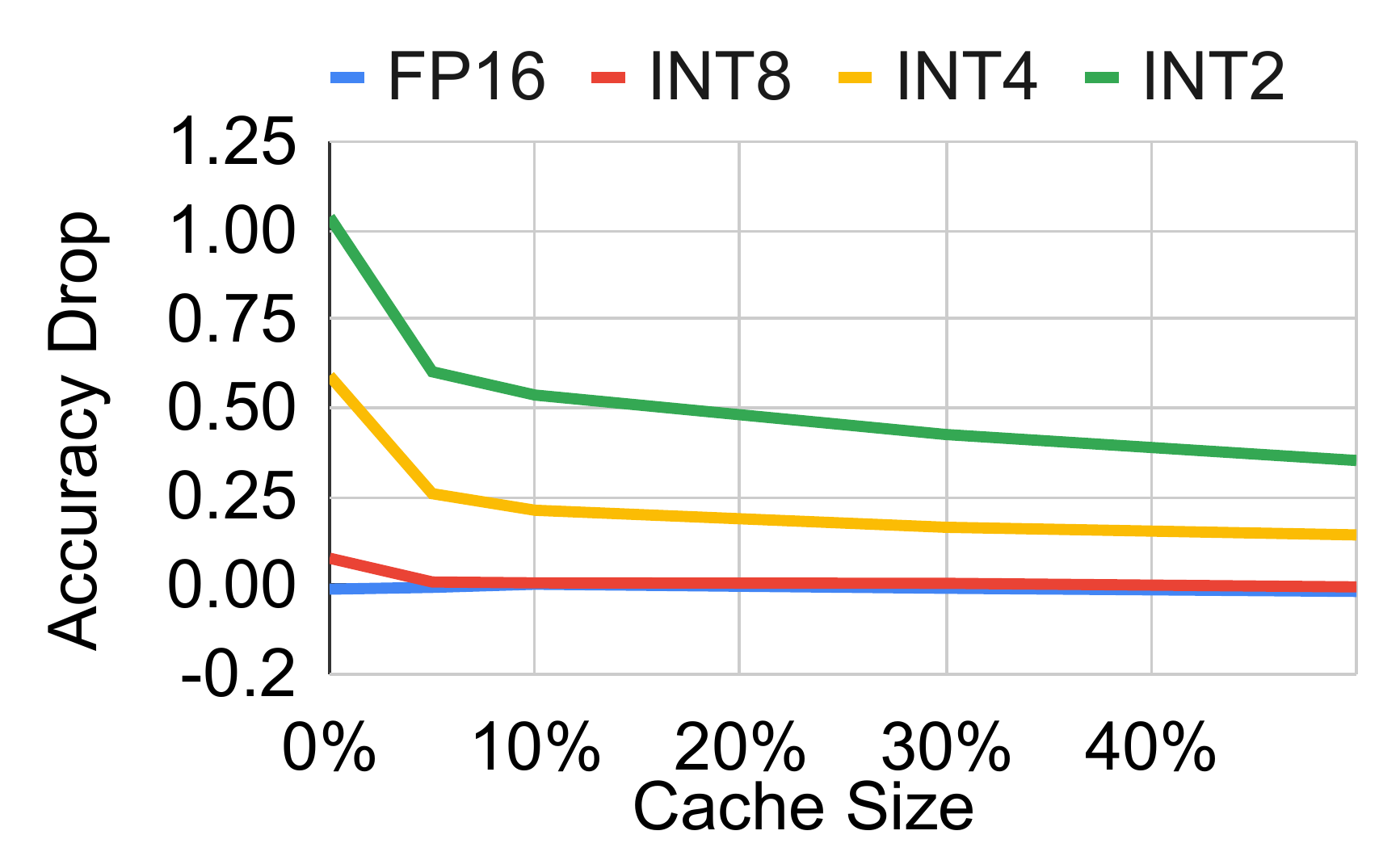} \\
   (a) LRU nearest rounding & (b) LRU stochastic rounding \\[6pt]
 \includegraphics[width=40mm]{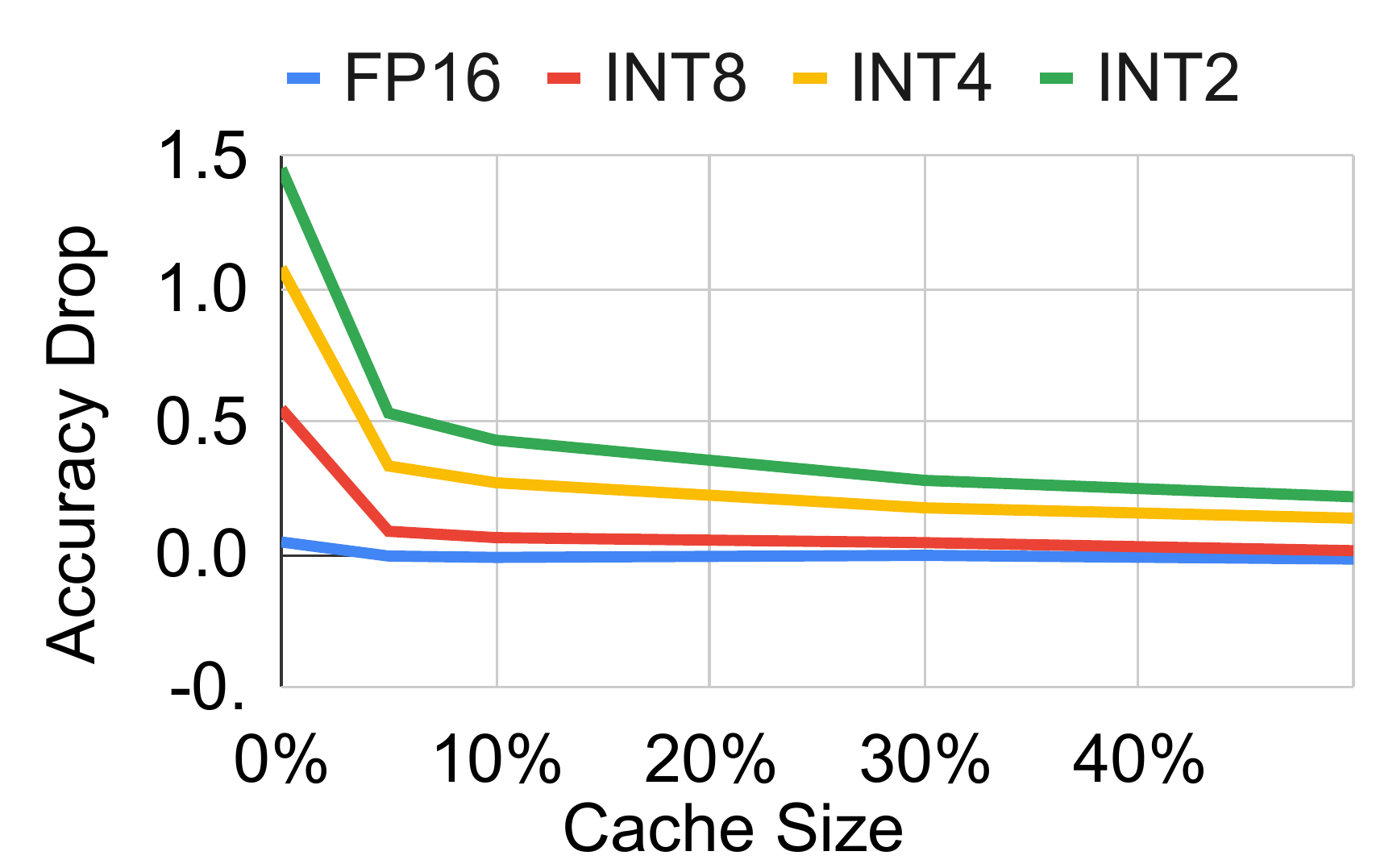} &  \includegraphics[width=40mm]{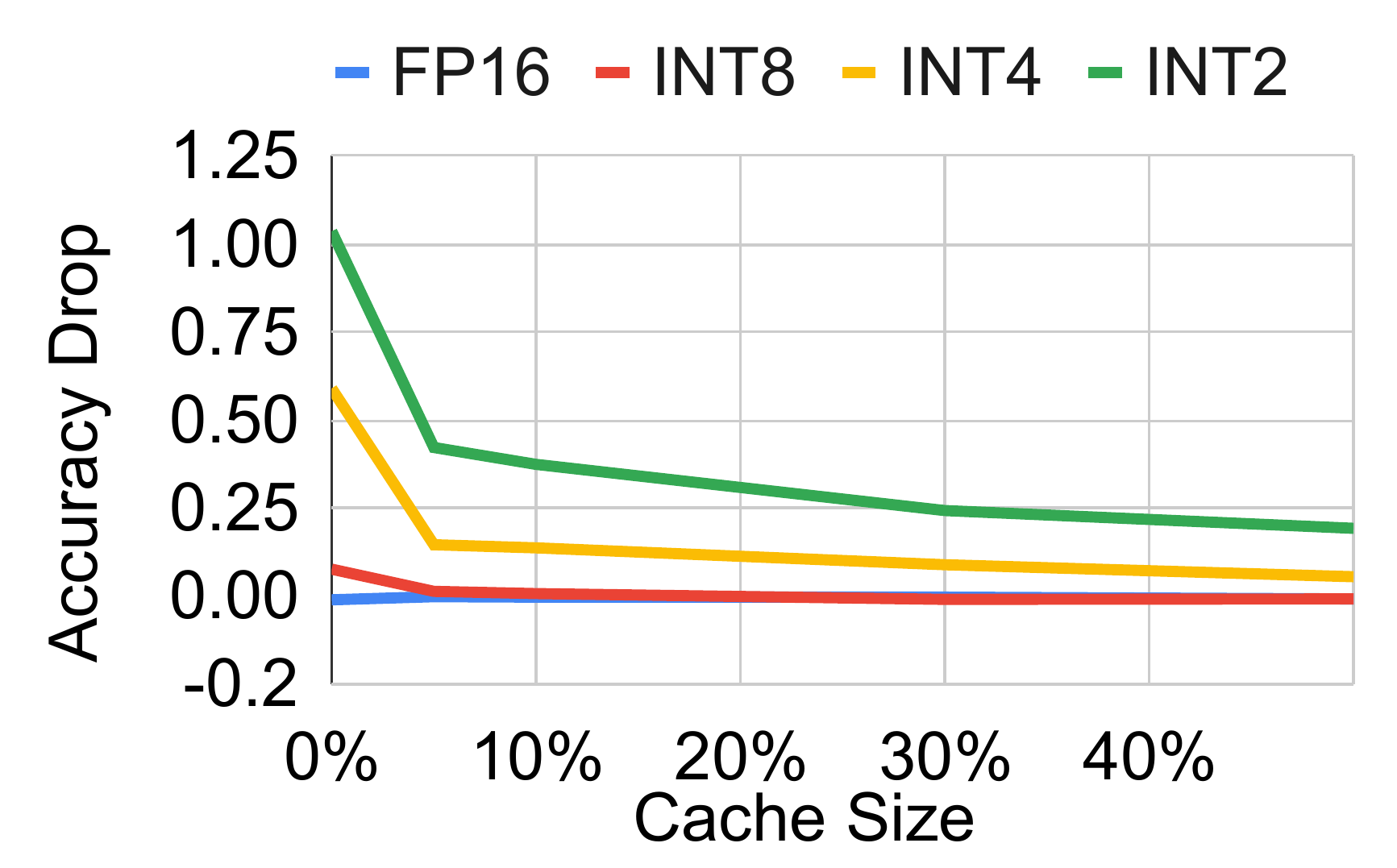} \\
(c) LFU nearest rounding & (d) LFU stochastic rounding \\[6pt]
\end{tabular}
\caption{Diminishing return of accuracy recovered vs. cache sizes}
\label{fig:cpu-cache-size}
\end{figure}

Although larger cache sizes recover more accuracy in
Tables \ref{tab:cpu_direct_lru} and \ref{tab:cpu_direct_lfu}, we observe a diminishing return of value in their continued increase.
Plotting in Figure \ref{fig:cpu-cache-size} the 16 columns of data in Tables \ref{tab:cpu_direct_lru} and \ref{tab:cpu_direct_lfu} demonstrate the trend.

\subsubsection{Replacement policy}\label{s:replacement_policy}

The previous section shows that the presence of a high-precision cache recovers significantly the accuracy lost due to embeddings trained exclusively in low precisions. When an FP32 cache is present, rows updated while in cache are computed in full FP32 precision. Thus any cache policy that promotes more updates to cache residents improve accuracy. This motivates us to examine the comparative number of updates each row receives during training. 

\begin{figure}
\includegraphics[width=85mm]{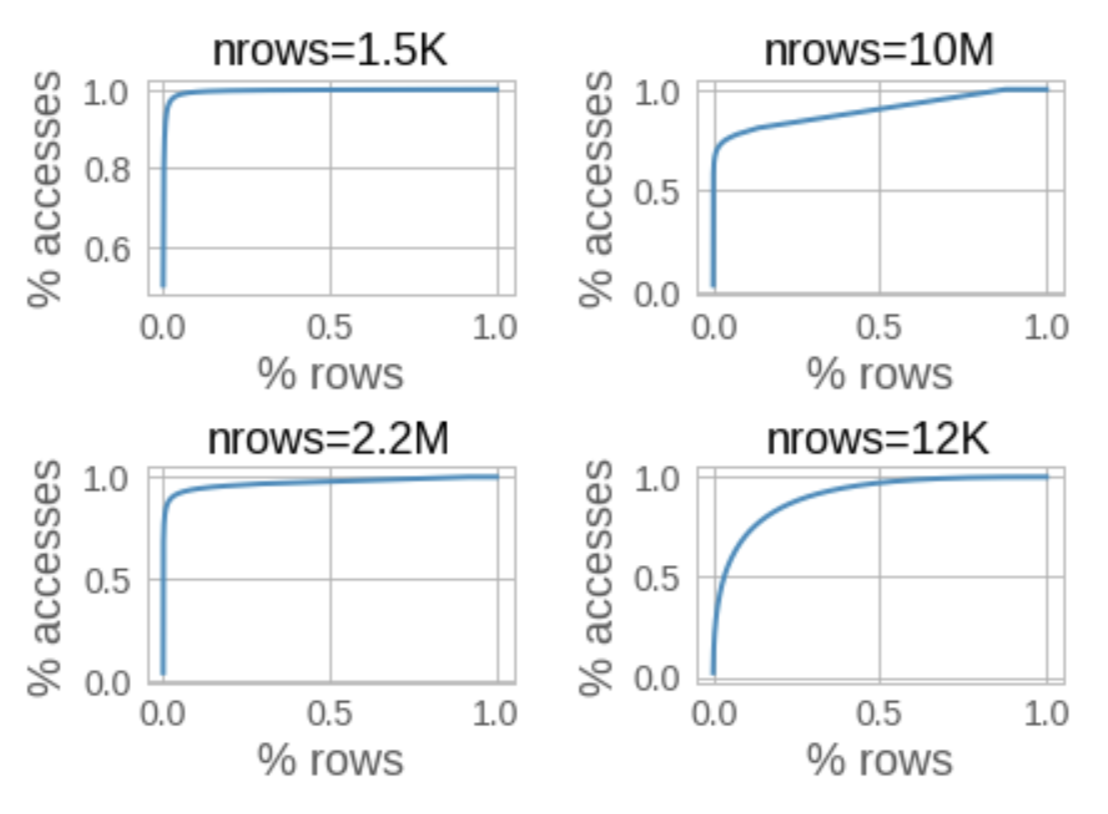}
\caption{CDF of sorted row access counts vs. fraction of rows}
\label{fig:cpu-row-cnts}
\end{figure}

We collected access counts for each row, and plotted cumulative distribution of row counters sorted in descending order. Figure \ref{fig:cpu-row-cnts} shows four of the representative distribution plots out of 15 embeddings with more than 1K rows. For Criteo-Kaggle dataset, all large embedding tables share the property that a small fraction ($<20\%$) of rows are responsible for a large fraction ($>80\%$) of total accesses; some embedding tables show an even greater skew. 

We now consider the relative merits of different cache replacement policies in light of the inequalities in access frequency. If the access pattern that produces these access count is non-stationary \cite{agarwal89} and highly phased, a direct-mapped LRU cache can be effective in promoting cache hit rate. An advantage is that direct-mapped LRU is the easiest to implement among all the choices in Algorithm \ref{alg:cpu-assoc}. On the other hand, if the access pattern is stationary, then the relative priority rankings among the rows based on access counts will more or less stabilize after some initial stages of training. Consequently, an LFU policy in general promotes the highly accessed rows be cache resident, increasing the number of cache hit. Along this line, an associative cache, with LFU policy in our case, is well known to improve cache hit rate further. 

\begin{table}
\caption{Cache hit rates for varying replacement policies and sizes (relative to embedding tables)}
\label{tab:cpu-cache-stats}
\vskip 0.15in
\begin{center}
\begin{small}
\begin{sc}
\begin{tabular}{rccc}
\toprule
cache size   &   LRU  & LFU  & 32-way LFU \\
\midrule
5\%  &  68.22\%   &   75.65\%  & 77.66\% \\ 
10\%  & 75.18\% & 81.21\% & 83.63\%   \\
30\% & 85.66\% & 89.07\% & 92.45\%\\
50\% & 90.12\% & 92.43\% & 96.13\%\\
\bottomrule
\end{tabular}
\end{sc}
\end{small}
\end{center}
\vskip -0.1in
\end{table}

\begin{figure}[h!]
\begin{tabular}{cc}
  \includegraphics[width=40mm]{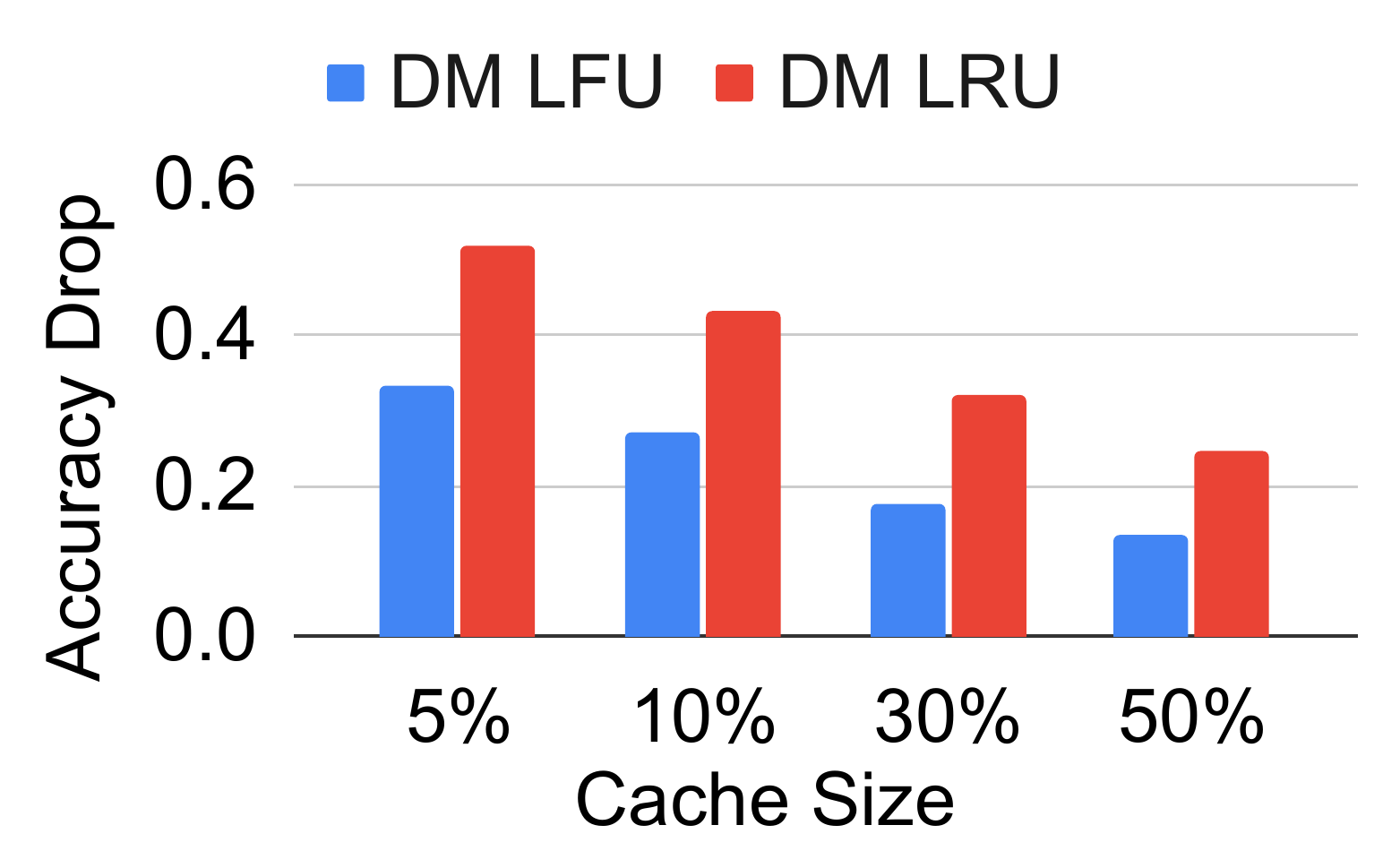} &   \includegraphics[width=40mm]{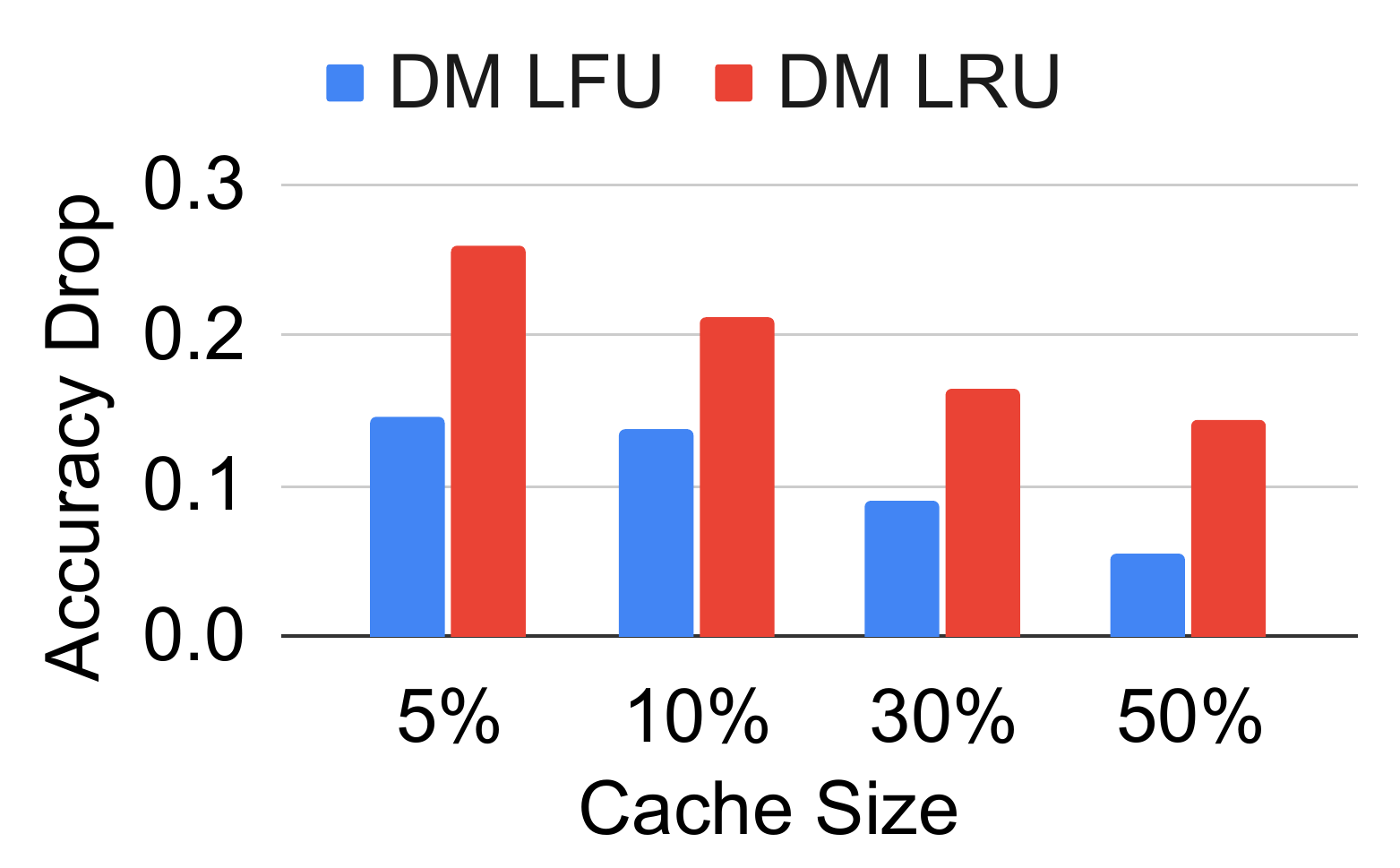} \\
   (a) INT4 nearest rounding & (b) INT4 stochastic rounding \\[6pt]
 \includegraphics[width=40mm]{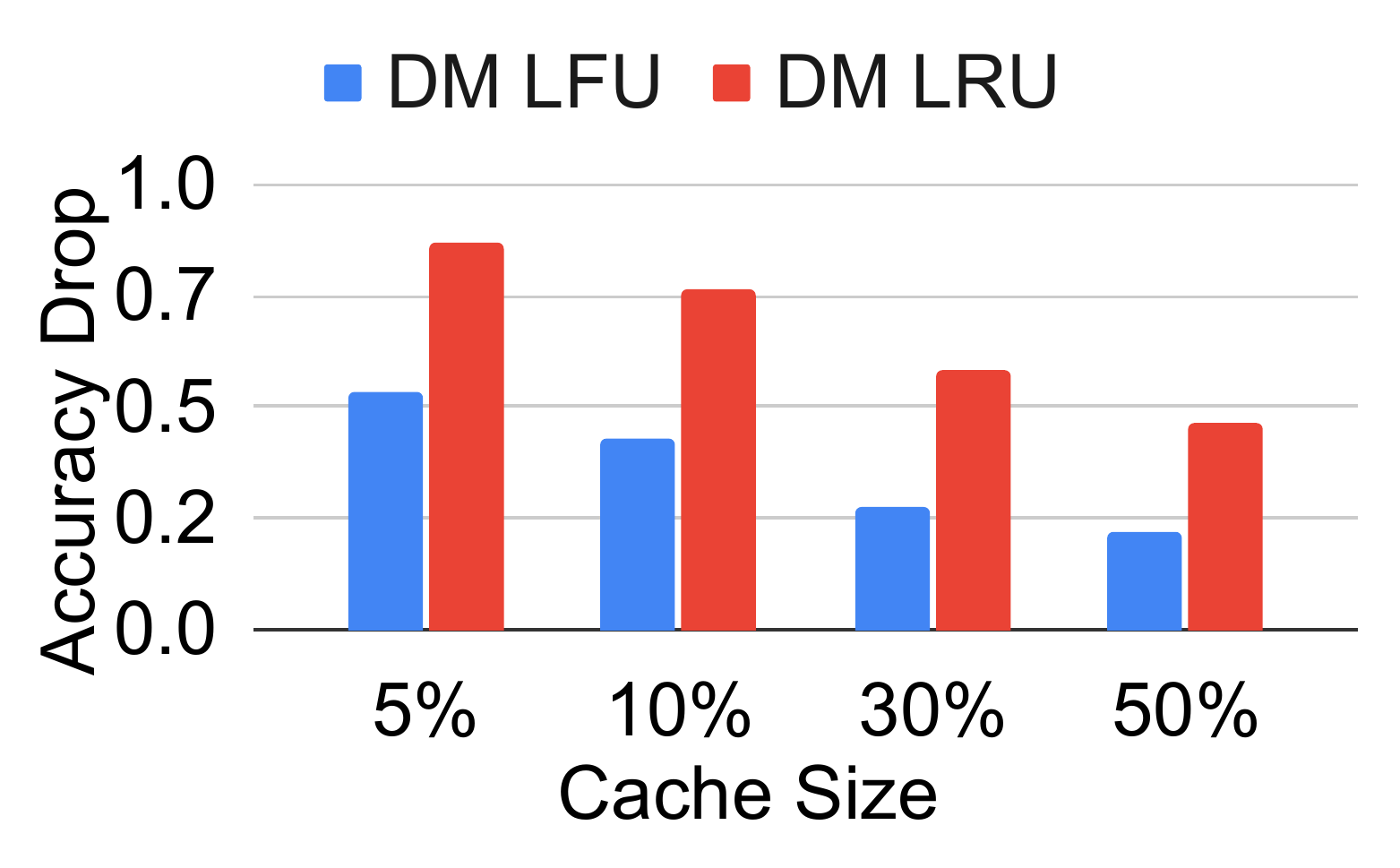} &  \includegraphics[width=40mm]{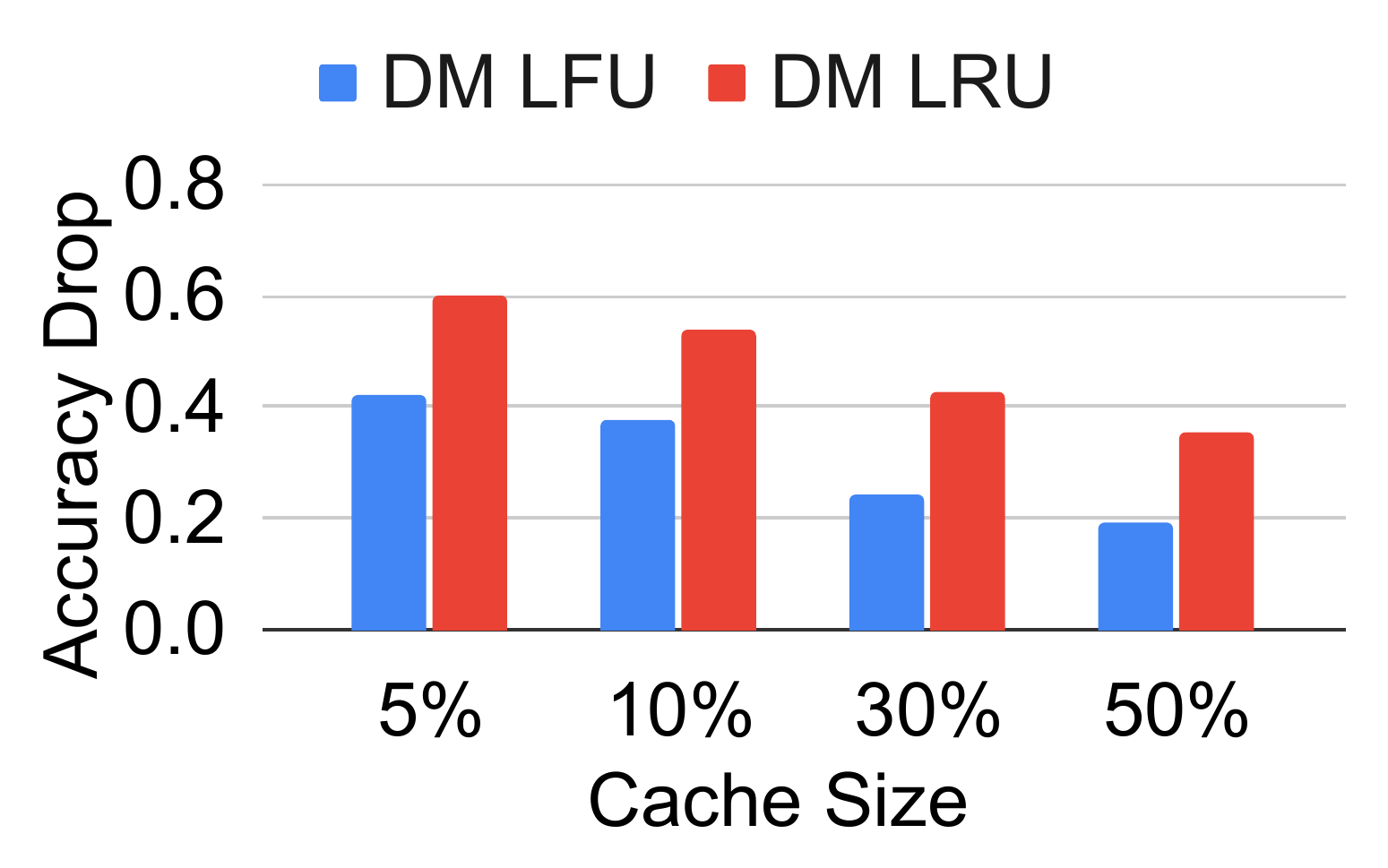} \\
(c) INT2 nearest rounding & (d) INT2 stochastic rounding \\[6pt]
\end{tabular}
\caption{Direct-mapped LFU vs. LRU}
\label{fig:cpu-lfu-v-lru}
\end{figure}

\begin{figure}[h!]
\begin{tabular}{cc}
  \includegraphics[width=40mm]{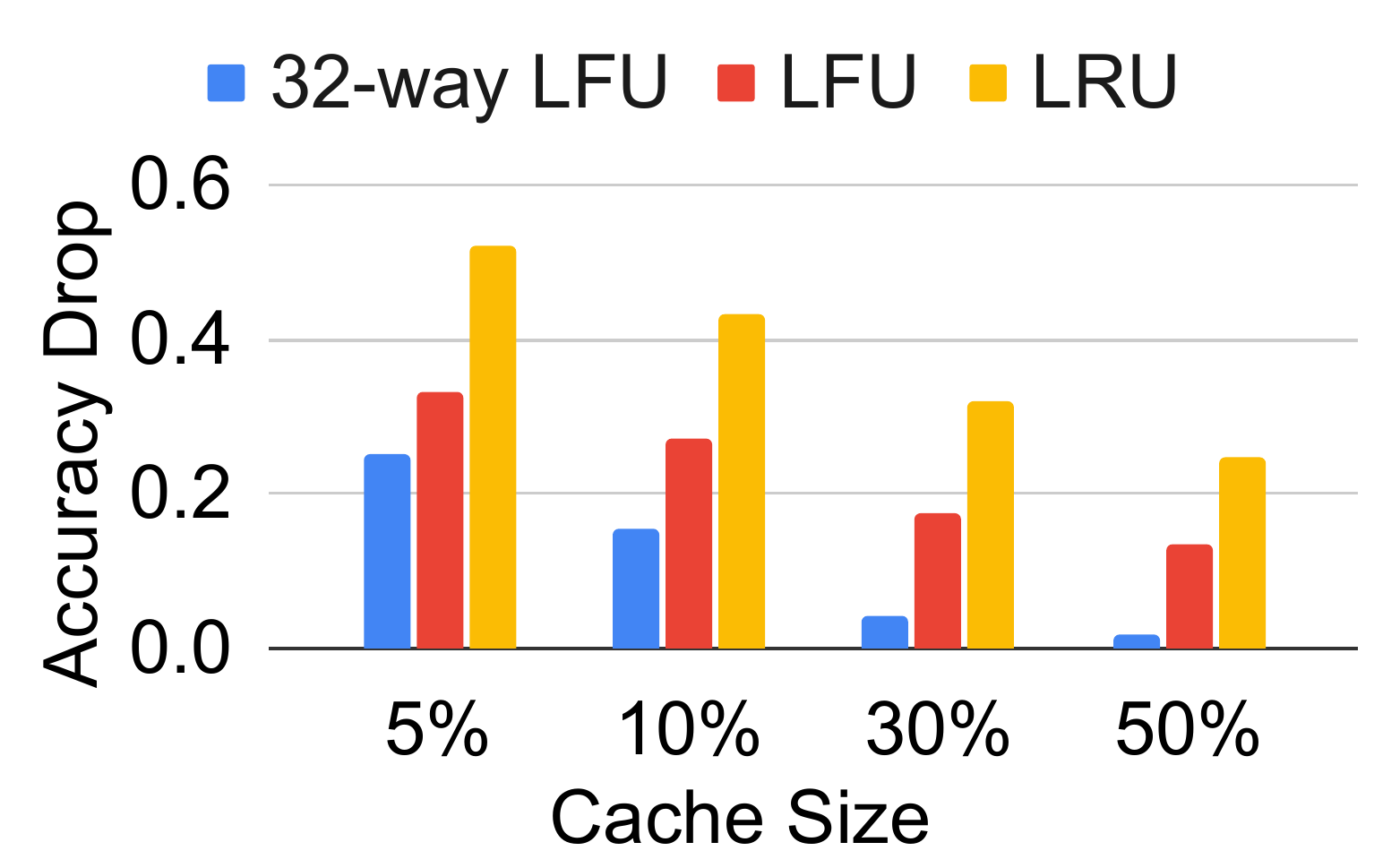} &   \includegraphics[width=40mm]{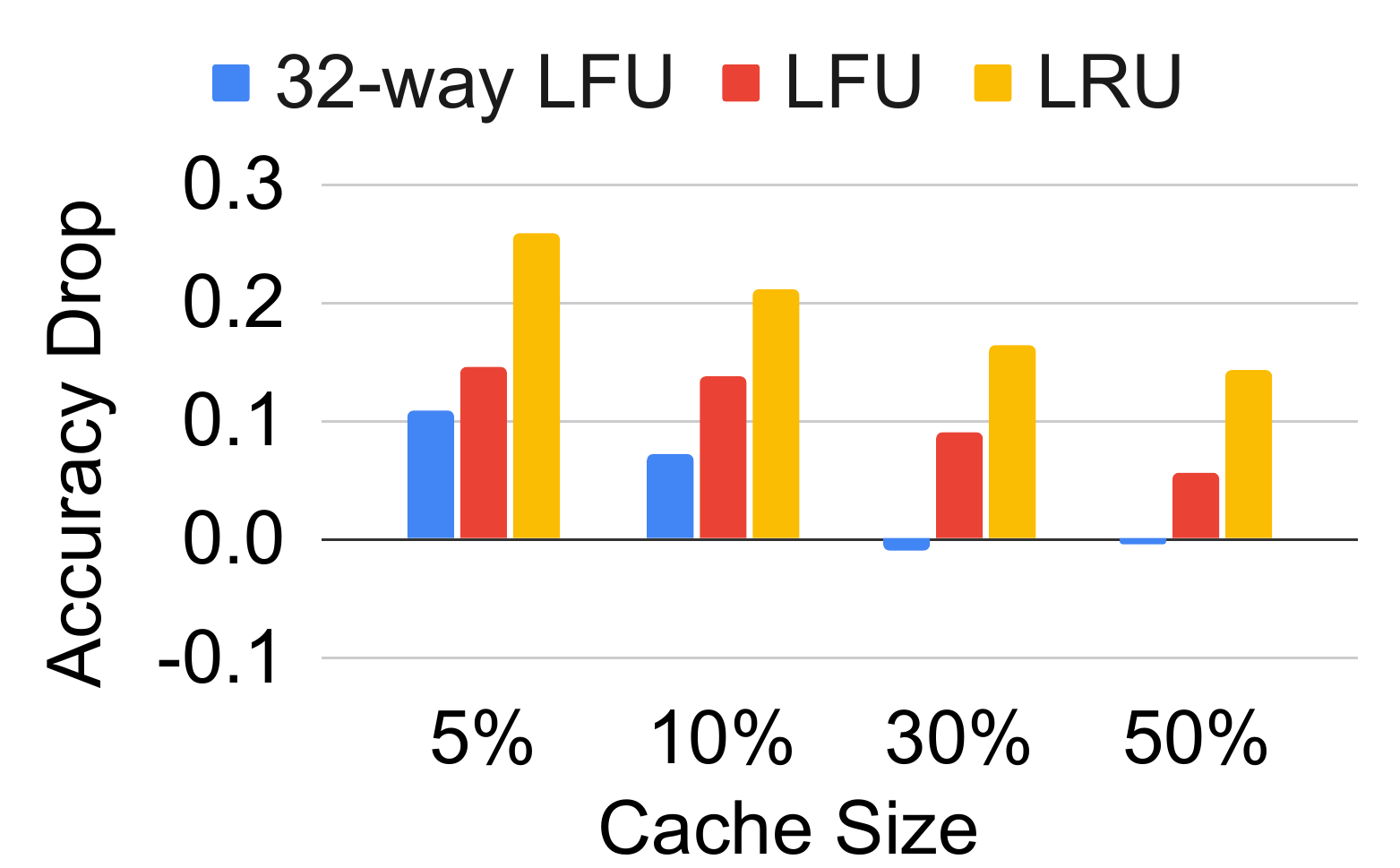} \\
   {\small (a) INT4 nearest rounding} & {\small (b) INT4 stochastic rounding} \\[6pt]
 \includegraphics[width=40mm]{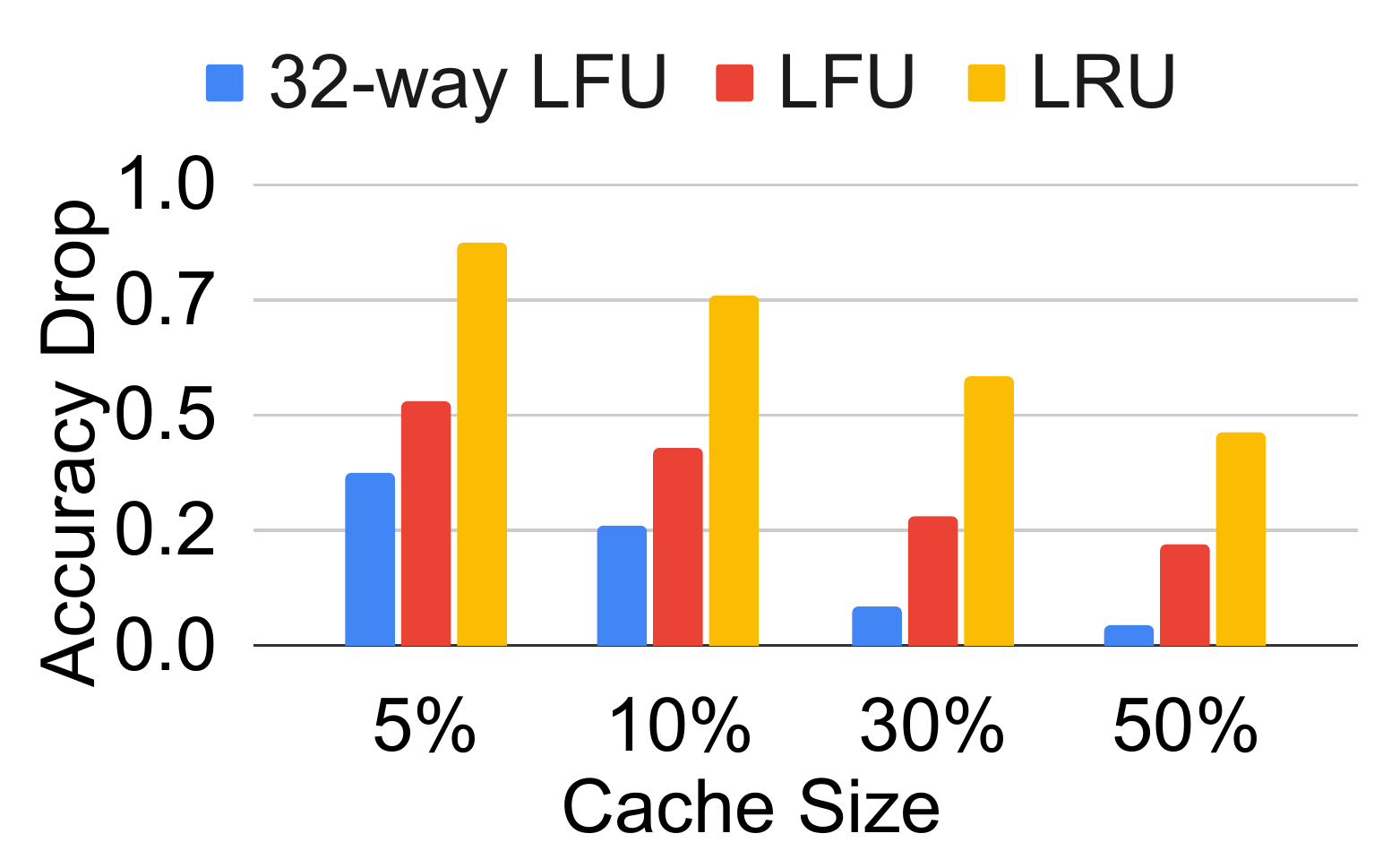} &  \includegraphics[width=40mm]{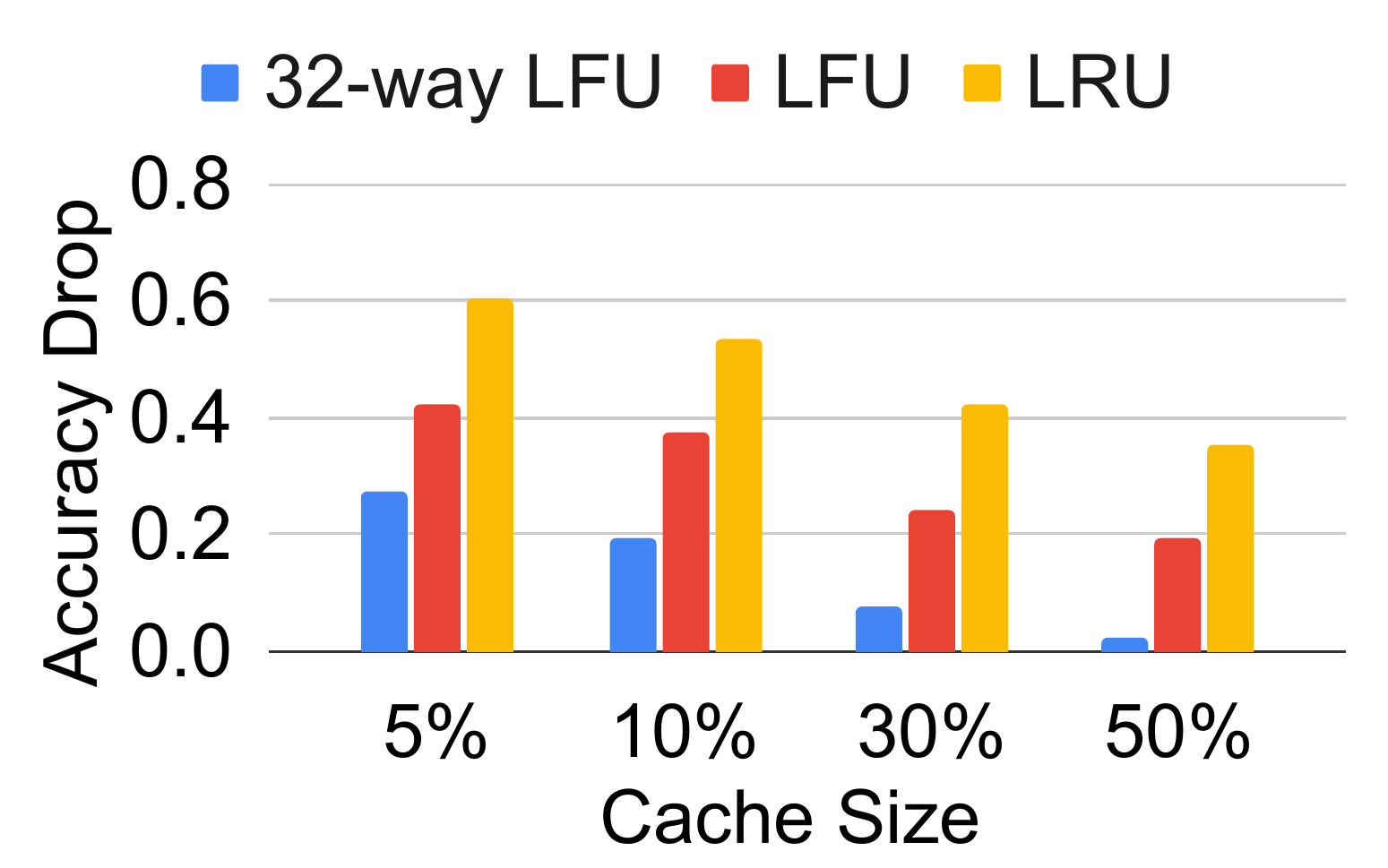} \\
{\small (c) INT2 nearest rounding} & {\small (d) INT2 stochastic rounding} \\[6pt]
\end{tabular}
\caption{32-Way Associative LFU vs. Direct-mapped LFU vs. LRU}
\label{fig:cpu-assoc-lfu-v-lru}
\end{figure}

\begin{figure}[h!]
\begin{tabular}{cc}
  \includegraphics[width=39mm]{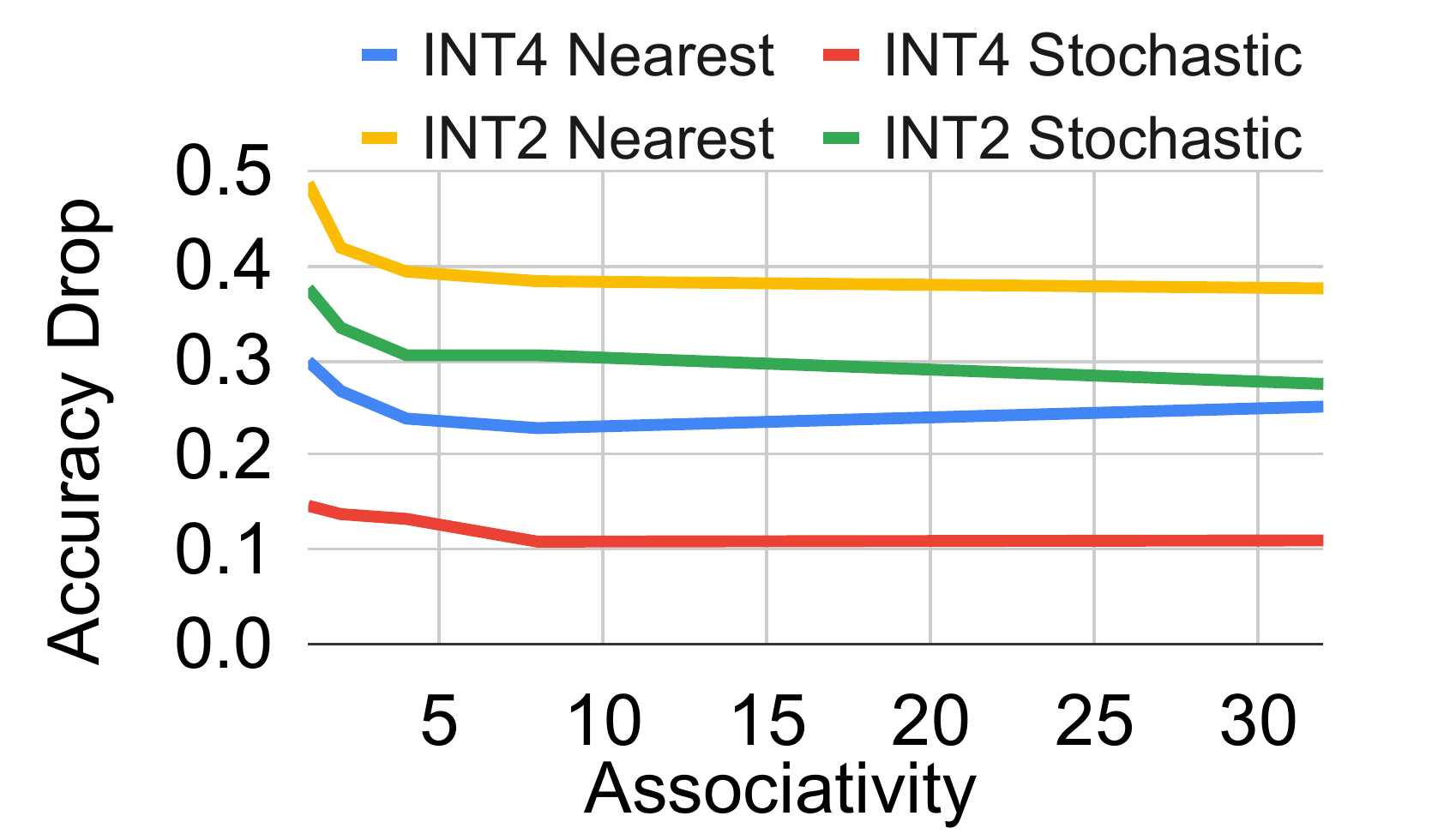} &   \includegraphics[width=39mm]{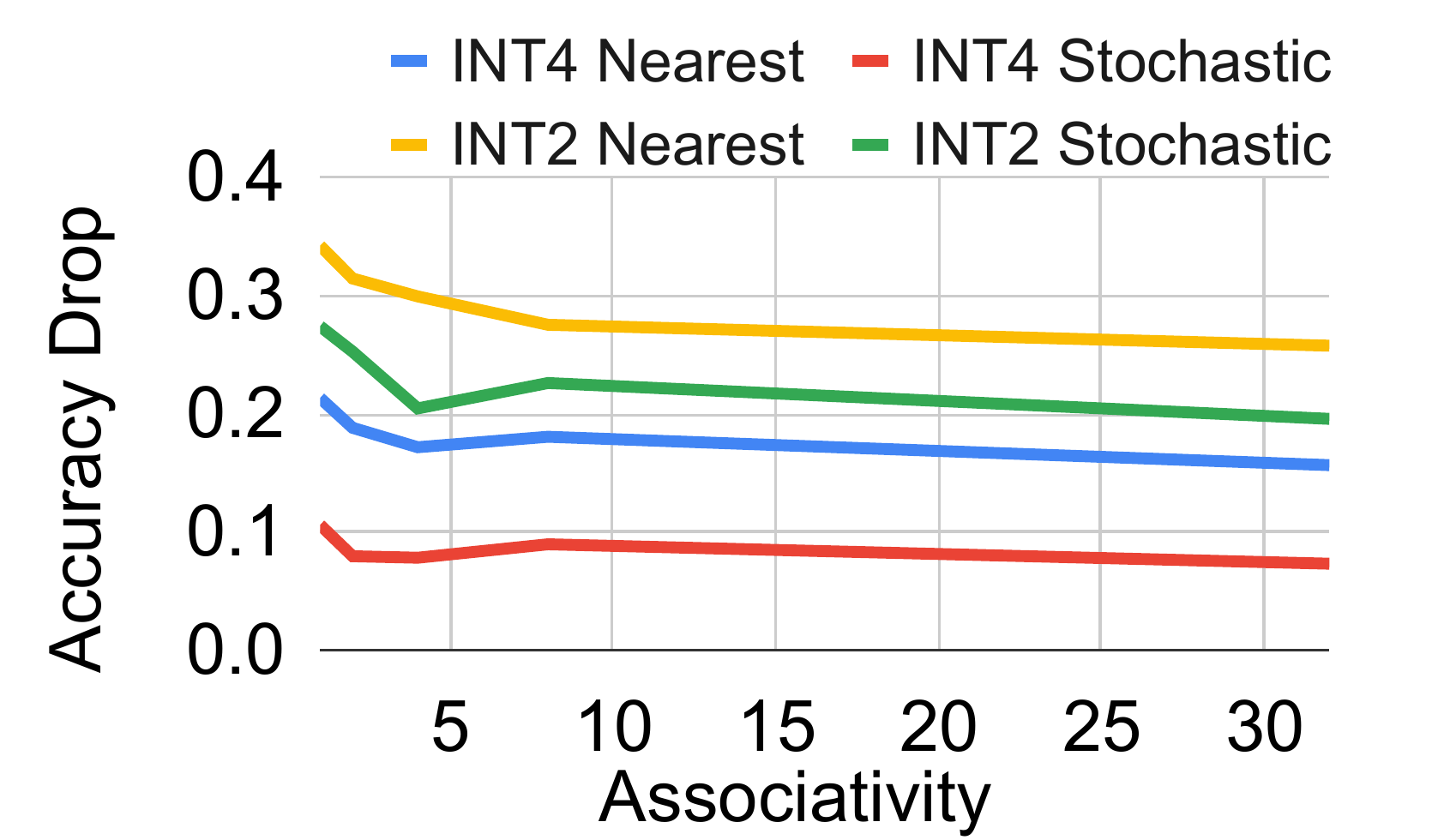} \\
   {\small (a) 5\% cache} & (b) {\small 10\% cache} \\[6pt]
 \includegraphics[width=39mm]{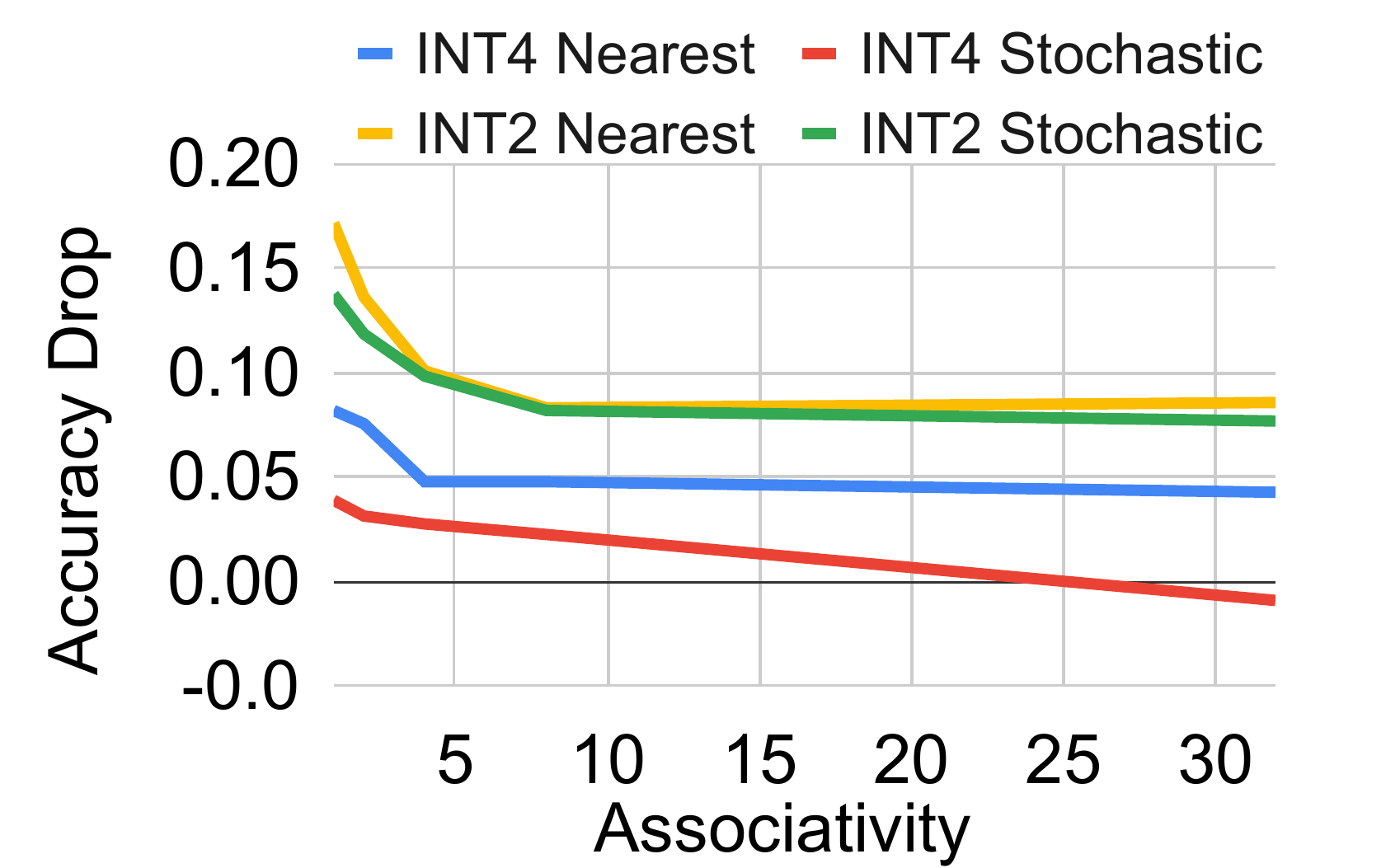} &  \includegraphics[width=39mm]{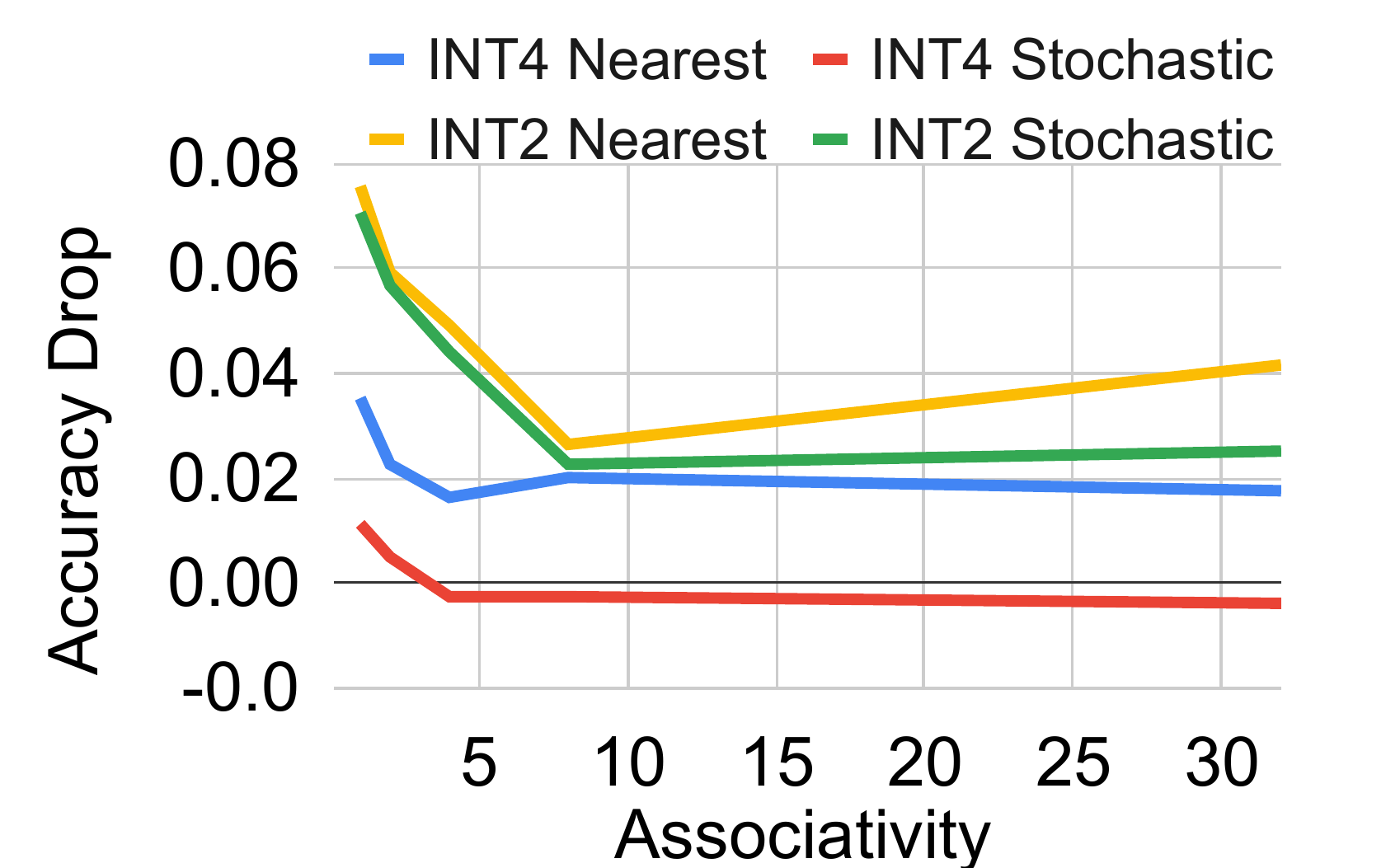} \\
{\small (c) 30\% cache} & {\small (d) 50\% cache} \\[6pt]
\end{tabular}
\caption{Diminishing return with associativity for set associative LFU cache}
\label{fig:assoc-lfu}
\end{figure}

Table \ref{tab:cpu-cache-stats} reports the cache hit rate statistics of various cache configurations. 32-way LFU provides noticeable increase in hit rate than direct-mapped LFU, while direct-mapped LFU has higher hit rates than direct-mapped LRU. These statistics are consistent with the hypothesis that the current model's embedding access pattern is somewhat stationary on the data set in question, making LFU and in particular with set-associative highly effective. The resulting model accuracy is well correlated positively with hit rate, as shown in Figures \ref{fig:cpu-lfu-v-lru} and \ref{fig:cpu-assoc-lfu-v-lru}. Figure \ref{fig:cpu-lfu-v-lru} shows that LFU outperforms LRU when both are direct mapped; Figure \ref{fig:cpu-assoc-lfu-v-lru} shows the superiority of set-associative LFU. That said, Figure \ref{fig:assoc-lfu} shows that accuracy gain from increased set associativity plateaus after a while, succumbing to the law of diminishing return.

\subsubsection{Results Summary}\label{s:cpu_summary}
We summarize our CPU experiments with DLRM and Criteo-Kaggle dataset. Most importantly, the use of a low-precision embedding table in conjunction with a high-precision cache is effective for maintaining accuracy as we have demonstrated. Stochastic rounding helps maintain training accuracy on low-precision embedding tables. We also see that set associative cache architecture is preferable to direct mapped, and that the LFU replacement policy serves this model/dataset better than LRU.

\begin{figure}[h!]
    \centering
    \includegraphics[width=3.2in]{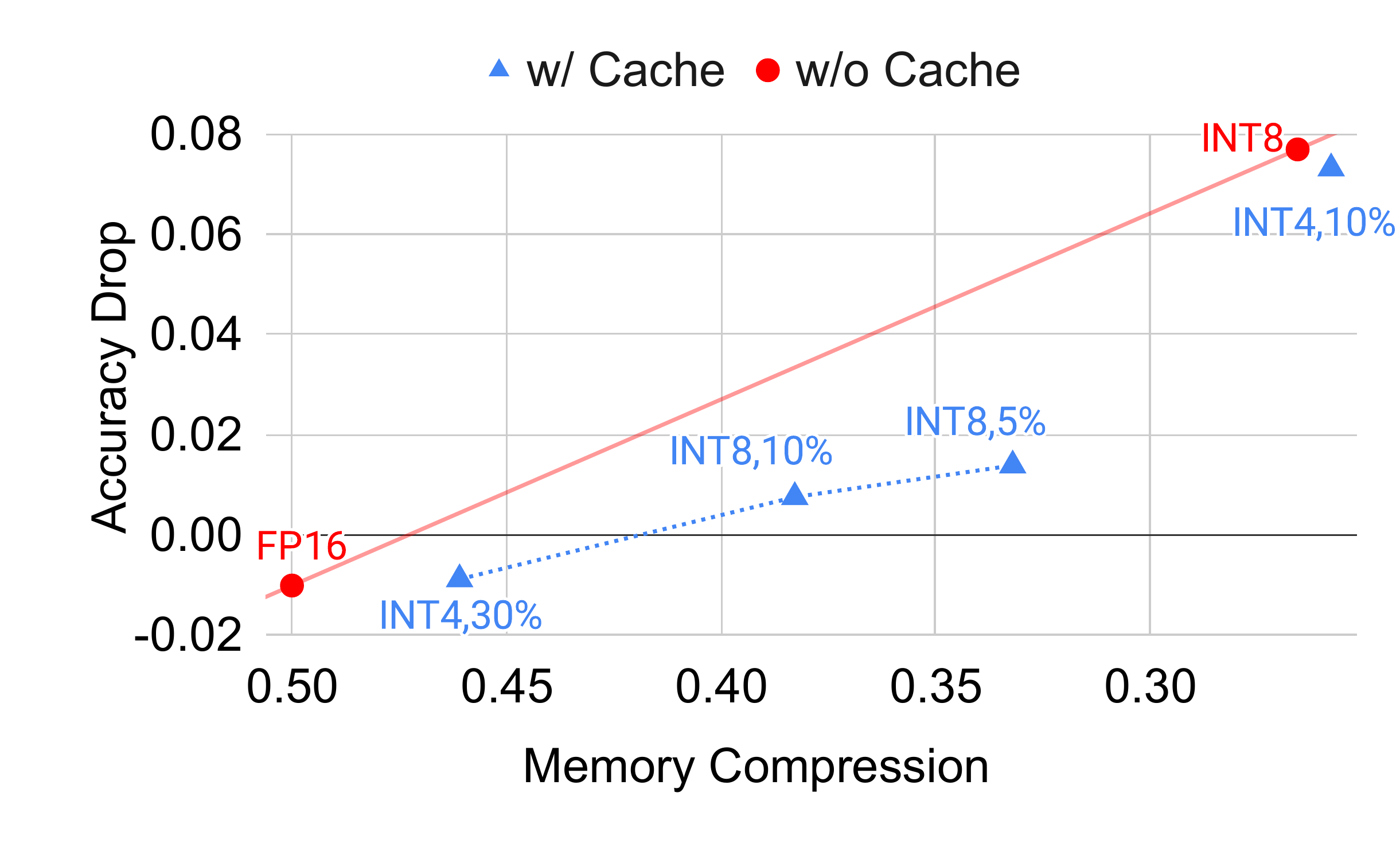}
    \caption{Accuracy vs. memory compression factor with and without high-precision cache on Criteo-Kaggle dataset. Please refer to Figure~\ref{fig:cpu-acc-reduction-complete} in Appendix~\ref{appendix:memory_compression_factor} for the complete figure.}
    \label{fig:cpu-acc-reduction}
\end{figure}

Concentrating then on a 32-way associative LFU cache with stochastic rounding, Figure~\ref{fig:cpu-acc-reduction} plots the best accuracy obtained at each memory compression factor with or without a high-precision cache. Memory compression factor with non-cached low-precision embedding is calculated from bitwidths, taking into quantization parameter storage overhead, i.e., compression for FP16 and INT8 is 0.5 and 0.265, respectively. Compression factor with cached implementation is calculated from bitwidths and cache sizes, taking into account quantization parameter storage overhead and caching overhead such as access counters and tags: i.e., for INT8 embedding with 10\% high-precision LFU cache, its memory compression factor is 0.374, or a 2.7$\times$ memory compression. Please refer to Appendix~\ref{appendix:memory_compression_factor} for more details for computing the memory compression factor.

Model accuracy degrades with increasingly large compression factors introduced by low-precision embeddings regardless of the presence of a high-precision cache. However with the addition of a high-precision cache, we effectively enable fine-grained tradeoffs between accuracy and memory footprint compression via varying cache sizes and precisions. A sweet spot here is a 3$\times$ memory compression (INT8 and 5\% cache) while maintaining neutral accuracy. For clarity to demonstrate the accuracy improvement near 0.02\% that we typically consider neutral, Figure~\ref{fig:cpu-acc-reduction} plots parts of our results, although the trend demonstrated holds true for lower precision than plotted in the figure. Please refer to Appendix~\ref{appendix:memory_compression_factor} for memory compression figure with complete results that show a general trend extending into lower precision.

\section{GPU Experiments with an Industry-scale Model and Dataset}\label{sec:gpu}
A GPU implementation enables us to validate our approach on larger datasets. We match the features of the CPU experimentation: supporting real FP16 and emulations of INT8/4/2 with rounding-to-nearest and stochastic rounding modes.

\subsection{Implementations}
We made the following changes to fully utilize the GPU architecture.
\begin{itemize}
\item {\bf 32-way only:} We implemented 32-way set-associative cache for: 1) our CPU results showed accuracy improvements over direct-mapped cache and 2) it matches well with the GPU's warp size of 32, hence achieving training speed similar to that of the direct-mapped case.
\item {\bf De-duplicate gradients:} Unlike the Kaggle dataset used for CPU evaluation with one-hot embedding accesses, the large dataset used for GPU evaluation has multi-hot embedding accesses. Therefore, there is a higher chance that multiple examples in a mini-batch access the same rows during sparse optimizer, leading to concurrent update races especially in GPUs with highly parallel execution that typically requires a large batch size. We avoid such races by sorting row indices in each mini-batch and merging gradient updates for the same rows into one update. Suppose one input in the mini-batch accesses rows 1 and 2 with gradient $g_1$, and another input accesses rows 2 and 3 with gradient $g_2$. We sort by the rows, de-duplicate gradients (e.g., $g_1 + g_2$ for row 2), and then apply the dedup'ed gradients.
\end{itemize}

\subsection{Experiments and Results}

To validate the feasibility of a high-precision cache on a larger model with larger dataset,
we evaluate an industrial scale DLRM-like recommendation model
for one ranking application at [institution name removed for double-blind review]
on an internal training dataset.
The size of the model and dataset are more than 10$\times$ bigger than the DLRM model and Kaggle dataset evaluated in the previous section, respectively.
Model accuracy is measured by 
Normalized-Entropy (NE)~\cite{ne_ref} metric. 
Lower NE corresponds to higher prediction accuracy for a recommendation model.
Similar to the DLRM evaluation on CPUs, we adopt relative accuracy drop in Equation~\ref{eq:accuracy_drop},
with the goal to keep the threshold below 0.02\%.

\subsubsection{Accuracy evaluation}

\begin{table}[h]
\caption{Accuracy Drop without Cache in \%. Note that "N/A" denotes the training cannot converge, resulting not a number (NaN) accuracy.}
\label{tab:gpu-no-cache-drop}
\begin{center}
\begin{small}
\begin{sc}
\begin{tabular}{lcccr}
\toprule
~   &   fp16 & int8 &  int4   &  int2    \\ 
\midrule
nearest  &  {\bf 0.001}   &   N/A & N/A  &  N/A \\ 
stochastic  & {\bf -0.023} & 4.254  &  5.200  & 4.673 \\
\bottomrule
\end{tabular}
\end{sc}
\end{small}
\end{center}
\vskip -0.1in
\end{table}

Table~\ref{tab:gpu-no-cache-drop} shows the accuracy drop without cache. Similar to the CPU experiments, FP16 can achieve neutral accuracy. However, accuracy drops more steeply with narrower bit-width of INT8/4/2, and doesn't even converge without stochastic rounding.


\begin{table}
\caption{High-precision LRU cache with varying cache sizes recover accuracy of low-precision embedding tables.}
\label{tab:gpu_lru_one}
\begin{center}
\begin{scriptsize}    
\begin{sc}
\begin{tabular}{l | c c c c }
\multicolumn{1}{c|}{
\makecell{Cache\\ Size\\ 
\% of}} &
\multicolumn{4}{c}{
\makecell{Accuracy Drop in \% \\ Cache: 32-way LRU on GPUs\\
Rounding: \nrnd{Nearest}/\srnd{Stochastic}}
} \\


\multicolumn{1}{c|}{Table} & FP16 & Int8 & Int4 & Int2 \\ \hline
0.1\% & \nrnd{\bf -0.010}/\srnd{\bf -0.024} & \nrnd{\bf 0.006}/\srnd{0.022} & \nrnd{0.053}/\srnd{\bf 0.010} & \nrnd{0.329}/\srnd{0.135} \\
1\% & \nrnd{\bf 0.004}/\srnd{\bf -0.027} & \nrnd{\bf -0.007}/\srnd{\bf 0.004} & \nrnd{\bf 0.002}/\srnd{\bf -0.001} & \nrnd{0.134}/\srnd{0.095} \\
 10\% & \nrnd{\bf -0.001}/\srnd{\bf -0.030} & \nrnd{\bf -0.011}/\srnd{\bf	0.014} & \nrnd{\bf -0.011}/\srnd{\bf -0.021} & \nrnd{0.024}/\srnd{0.028} \\
\end{tabular}
\end{sc}
\end{scriptsize}
\end{center}
\end{table}


\begin{table}[h!]
\caption{High-precision LFU cache with varying cache sizes recover accuracy of low-precision embedding tables}
\label{tab:gpu_lfu_one}
\begin{center}
\begin{scriptsize}    
\begin{sc}
\begin{tabular}{l | c c c c }
\multicolumn{1}{c|}{
\makecell{Cache\\ Size\\ 
\% of}} &
\multicolumn{4}{c}{
\makecell{Accuracy Drop in \% \\ Cache: 32-way LFU on GPUs\\
Rounding: \nrnd{Nearest}/\srnd{Stochastic}}
} \\
\multicolumn{1}{c|}{Table} & FP16 & Int8 & Int4 & Int2 \\ \hline


 0.1\% & \nrnd{\bf 0.002}/\srnd{\bf -0.029} & \nrnd{0.041}/\srnd{0.129} & \nrnd{0.112}/\srnd{0.133} &\nrnd{0.526}/\srnd{0.278} \\
 1\% & \nrnd{\bf 0.011}/\srnd{\bf -0.022} & \nrnd{0.038}/\srnd{0.081} & \nrnd{0.065}/\srnd{0.081} & \nrnd{0.389}/\srnd{0.189} \\
 10\% & \nrnd{\bf 0.003}/\srnd{\bf -0.017} & \nrnd{\bf -0.003}/\srnd{\bf 0.011} & \nrnd{\bf 0.002}/\srnd{\bf 0.009} & \nrnd{0.057}/\srnd{0.042} \\
\end{tabular}
\end{sc}
\end{scriptsize}
\end{center}
\end{table}


Then, we apply the 32-way high-precision LFU and LRU cache with different sizes: 0.1\%, 1\%, and 10\% of the original table size, for large embedding tables that collectively account for more than 97\% of the total model size. Tables~\ref{tab:gpu_lru_one} and \ref{tab:gpu_lfu_one} demonstrate that on the large training dataset, both LFU and LRU can achieve better accuracy for various precisions with nearest/stochastic rounding.
Unlike the results with DLRM and Kaggle dataset, LRU performs generally better than LFU. This will be explained in the next section.

\subsubsection{Cache hit rate analysis}

\begin{table}[b!]
\caption{Cache hit rates for varying replacement policies and sizes (relative to embedding tables). Caching is applied to two sets of tables. The first set is 5\% of embedding tables accounting for 97.1\% of total model size.}
\label{tab:gpu-cache-stats}
\vskip 0.15in
\begin{center}
\begin{small}
\begin{sc}
\begin{tabular}{ccccr}
\toprule
  \makecell{Cache}           &   \multicolumn{2}{c}{5\% of tables}  & \multicolumn{2}{c}{55\% of tables}  \\
  \makecell{Size} & \multicolumn{2}{c}{97.1\% of size} & \multicolumn{2}{c}{99.9\% of size} \\
  \makecell{in \%}  &   LRU  & LFU  & LRU & LFU \\
\midrule
0.1\% & 10.52\%	& 2.41\% & 33.50\% & 74.97\% \\
1\%  & 34.72\% & 13.41\% & 94.06\% & 96.11\% \\
10\% & 59.86\% &	57.31\% &	99.96\% &	99.95\% \\
\bottomrule
\end{tabular}
\end{sc}
\end{small}
\end{center}
\vskip -0.1in
\end{table}


To understand the data reuse pattern of the large training dataset, we collected the cache hit rate statistics for different cache sizes and replacement policies (Table~\ref{tab:gpu-cache-stats}).
Contrary to the Criteo-Kaggle 7D Ads display challenge dataset, when we apply caching to 5\% of tables that account for 97\% of total model size, LRU performs consistently better than LFU for the internal training dataset, especially for smaller cache sizes.
However, if we apply caching to more tables, LFU performs better than LRU.
One reason for this difference is the ``data shift'' pattern in the internal dataset for a few large tables, where some entity ids are popular for a continued period of time exhibiting temporal locality but later don't appear as frequently.




\subsubsection{Memory Reduction}

\begin{figure}
    \centering
    \includegraphics[width=3.3in]{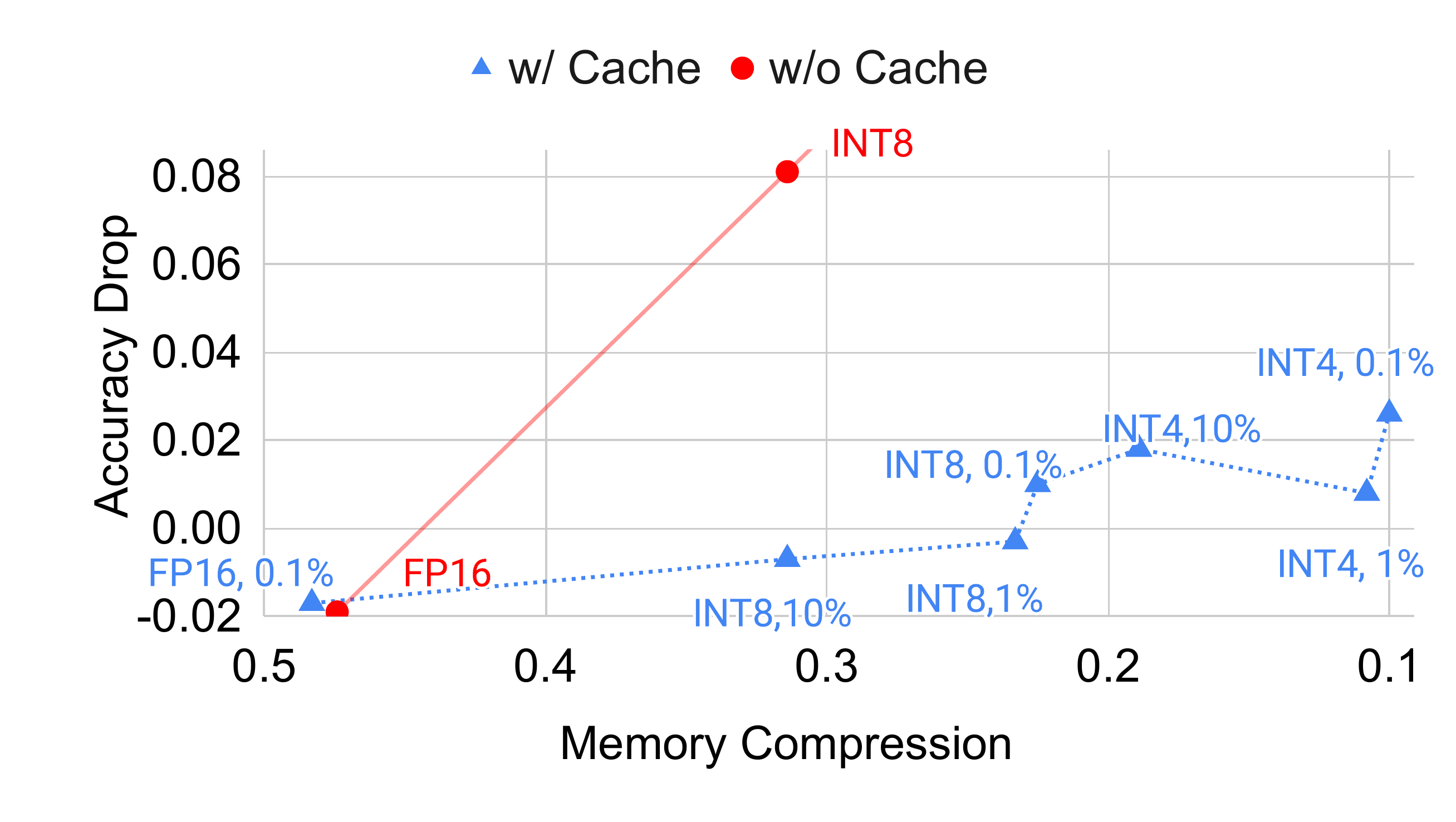}
    \caption{Accuracy vs. memory compression factor with and without high-precision cache for large training dataset.}
    \label{fig:gpu-acc-reduction}
\end{figure}

Similar to the analysis in Section~\ref{s:cpu_summary}, we calculate the memory compression factor for both non-cached and cached implementations of low-precision embeddings on GPUs, taking into account quantization parameter storage overhead and cache-specific access counters as well as tags. For example, the memory compression factor for INT8 embedding with 1\% high-precision LRU cache is 0.259. Note that LFU takes a bit more storage than LRU implementations since LFU needs to maintain the extra counters to record the frequency.

Figure~\ref{fig:gpu-acc-reduction} shows the correspondences between accuracy and memory reduction with and without cache. For clarity to demonstrate accuracy differences for the settings that give close to neutral accuracy, we only plot parts of our results in Figure~\ref{fig:gpu-acc-reduction}. The trend demonstrated holds true for lower precision that are not shown in the figure as well.

For example, with the accuracy drop threshold 0.02\%, we can use INT4 stochastic rounding embedding with 1\% high-precision cache with 86.6\% memory savings on GPUs (more than 7$\times$ memory reduction), which significantly mitigates the limited on-device high-bandwidth memory (HBM) capacity of GPUs.



\subsubsection{Speed improvement}


\begin{figure}[h!]
\begin{tabular}{cc}
  \includegraphics[width=78mm]{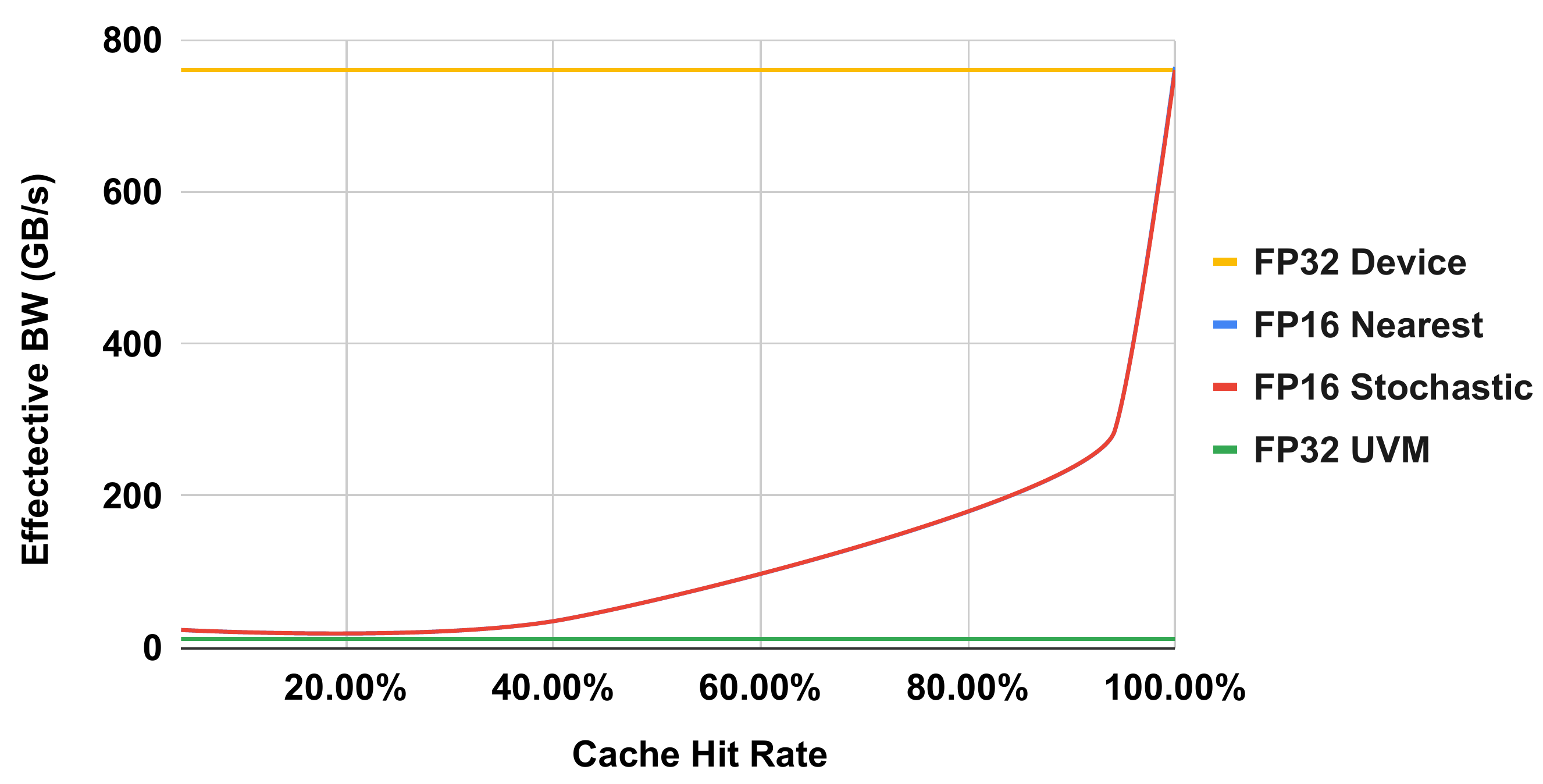} \\  
  {\small (a) LRU, forward. {FP16 Near./Stoc. \textbf{overlap} with each other.}} \\[6pt]\includegraphics[width=78mm]{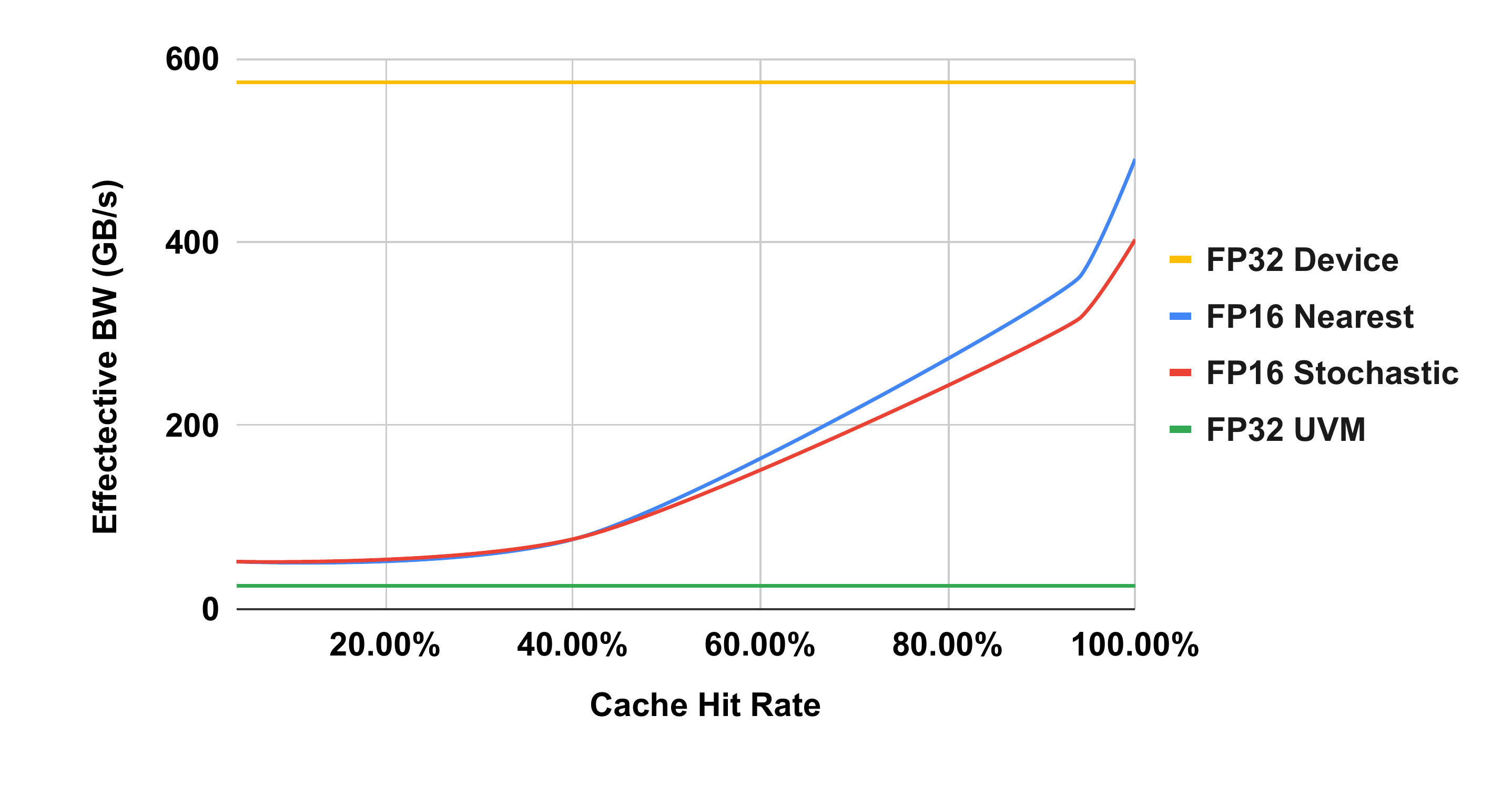} \\
   {\small (b) LRU, backward}
\end{tabular}
\caption{Effective bandwidth for FP32 (for embedding table on device and on unified memory) vs. FP16 with high-precision LRU cache. The effective bandwidth of LFU is similar.}
\label{fig:gpu-speed}
\end{figure}

When embedding tables are too big to fit on-device HBM memory of GPUs, one should use mechanisms like unified memory (UVM, \cite{unified_memory}. However, UVM operates in a granularity considerably bigger than embedding table rows with non-contiguous access pattern, significantly wasting PCIe bandwidth. By using low-precision embedding tables with high-precision cache, we can either (1) reduce the size of cache plus embedding table small enough to fit the both entirely within device memory, or (2) still put the low-precision embedding table in the unified memory but reducing PCIe traffic.

We want to make sure that our GPU cache implementation is fast enough to outperform UVM when cache hit rate is low, and saturates a large fraction of peak device memory bandwidth when hit rate is high. This is demonstrated by Figure~\ref{fig:gpu-speed}.

Figure~\ref{fig:gpu-speed} shows micro-benchmark results evaluated on a Nvidia V100 GPU, where a large low-precision embedding table is allocated on the unified memory (UVM) and a high-precision cache is allocated on the device memory. We plot the effective bandwidth (in GB/s) in correspondence to different cache hit rate for LRU cache during the forward and backward passes . The top and bottom lines show the achieved bandwidth of the forward/backward path on the device memory and on the unified memory, which correspond to the upper and lower bound of the bandwidth. Note that the bandwidth of the device memory is bounded by the HBM peak bandwidth (900 GB/s), while the bandwidth of the unified memory is bounded by the PCIe bandwidth. When the hit rate is high, we get close to the device bandwidth. For lower hit rate, the LRU/LFU implementation is still reasonably faster than the bandwidth achieved by unified memory.

We also measure an end-to-end training speedup of 16\% using FP16 LRU cache (cache size is 1\% of embedding tables) over the baseline where UVM is used for large embedding tables. This is an example that our technique not only saves memory but can also improve the training speed.


\section{Conclusions and Future Work}\label{sec:conclusion}

In this paper we presented low-precision embedding table with a high-precision cache as an effective memory saving technique for training large-scale recommendation models, particularly well suited for systems with high compute bandwidth but limited capacity memory like GPUs. To the best of our knowledge, this is the first time mixed-precision training with a high-precision cache is applied on an industrial-scale recommendation system. Our best results include reducing memory by over 7$\times$ for a large embedding trained in INT4 precision while maintaining neutral accuracy. 

As part of future work, we will further experiment with heterogeneous precision and replacement policy assignment for different embeddings. We will potentially assign different precision and cache policy for each embedding based on a light-weight profile in the first few training iterations for the best accuracy. This can be useful to handle dataset with different characteristics as we have seen from the Kaggle and our internal dataset. We will further explore different hash functions and replacement heuristics, and evaluate their impact on model accuracy and performance.
Our technique can be applied to other models with embeddings such as language models. Once proven to work for a wider range of models, it can be interesting to add hardware support by augmenting low-precision conversion to the existing hardware cache mechanisms.





\clearpage
\newpage

\balance

\bibliography{main}
\bibliographystyle{mlsys2020}


\clearpage
\newpage
\nobalance
\appendix
\section{Appendix: Memory Compression Factor}\label{appendix:memory_compression_factor}
%

\begin{table*}
\caption{Calculation of memory compression factor}
\label{appendix:compression}
\vskip 0.15in
\begin{center}
\begin{small}
\begin{sc}
\begin{tabular}{ccccccc}
\toprule
Bitwidth & Cache size  & Bits per row & Bits per row & Access counter  & Cache tags  &  Compression  \\

~ & ratio &  (FP32) & (quantized) & bits per row & bits per row & factor \\

\midrule
$ 8 $ & $ 0 $ & $ 128 \times 32 $    & $ 8 \times 128+64 $   & $ 0 $   & $ 0 $   & $ 0.26563 $ \\
$ 4 $ & $ 0 $ & $ 128 \times 32 $   & $ 4 \times 128+64 $    & $ 0 $   & $ 0 $  & $ 0.14063 $  \\
$ 2 $ & $ 0 $ & $ 128 \times 32 $   & $ 2\times 128+64 $   & $ 0 $  & $ 0 $  & $ 0.07813 $ \\
\midrule
$ 4 $ & $ 0.3  $ & $ 128 \times 32 $  & $ 4 \times 128+64 $ & $ 32 $  & $ 32 \times 0.3  $  & $ 0.45078 $ \\
$ 8 $ & $ 0.1  $ & $ 128 \times 32 $  & $ 8 \times 128+64 $ & $ 32 $  & $ 32 \times 0.1  $  & $ 0.37422 $ \\
$ 8 $ & $ 0.05 $ & $ 128 \times 32 $  & $ 8 \times 128+64 $ & $ 32 $  & $ 32 \times 0.05 $  & $ 0.32383 $ \\
$ 4 $ & $ 0.1  $ & $ 128 \times 32 $  & $ 4 \times 128+64 $ & $ 32 $  & $ 32 \times 0.1  $  & $ 0.24922 $ \\
$ 4 $ & $ 0.05 $ & $ 128 \times 32 $  & $ 4 \times 128+64 $ & $ 32 $  & $ 32 \times 0.05 $  & $ 0.19883 $ \\
$ 2 $ & $ 0.1  $ & $ 128 \times 32 $  & $ 2 \times 128+64 $ & $ 32 $  & $ 32 \times 0.1  $  & $ 0.18672 $ \\
$ 2 $ & $ 0.05 $ & $ 128 \times 32 $  & $ 2 \times 128+64 $ & $ 32 $  & $ 32 \times 0.05 $  & $ 0.13633 $ \\
\bottomrule
\end{tabular}
\end{sc}
\end{small}
\end{center}
\vskip -0.1in
\end{table*}

Memory compression factor is used to characterize the compressed model size after applying the low-precision embedding table with a high-precision cache. We need to consider the following dominant components:
\begin{itemize}
\item Low-precision table size, $ M_{T_{lowprec}} $;
\item Quantization parameter storage size, $ M_{Q} $;
\item High-precision cache size, $ M_{C} $;
\item Cache tags size, $ M_{tags} $;
\item Access counters (if using LFU), $ M_{cnt} $;
\item Original FP32 embedding table size, $ M_{T_{FP32}} $
\end{itemize}
Specifically, the ``memory compression factor'' is calculated as
\begin{equation}
    (M_{T_{lowprec}} + M_{C} + M_{tags} + M_{cnt}) / M_{T_{FP32}}
\end{equation}

where:
\begin{itemize} 
    \item Cache tags are INT32 numbers, one per each row in cache;
    \item Access counters are INT32 numbers, one per each row of the embedding.
\end{itemize}

Assume the number of embedding rows is $ E $, the embedding dimension is $ D $, and the low-precision bitwidth is $ bitwidth $, then the low-precision table size is calculated as:
\begin{equation}
M_{T_{lowprec}} = E \times (bitwidth \times D + 64),
\end{equation}
where $ 64 $ is the per-row overhead of quantization parameters (in FP32): 32 bits for the per-row scale and 32 bits for the per-row bias. 
 
Table~\ref{appendix:compression} shows memory compression factors calculated for a sample of precision and cache size combinations using a LFU cache, and the embedding dimension $ D = 128 $ (used in CPU experiments). As the number of rows per embedding is different for each table, we compute cache and quantization overhead as the number of bits per embedding row.

\begin{figure*}
    \centering
    \includegraphics[width=6.5in]{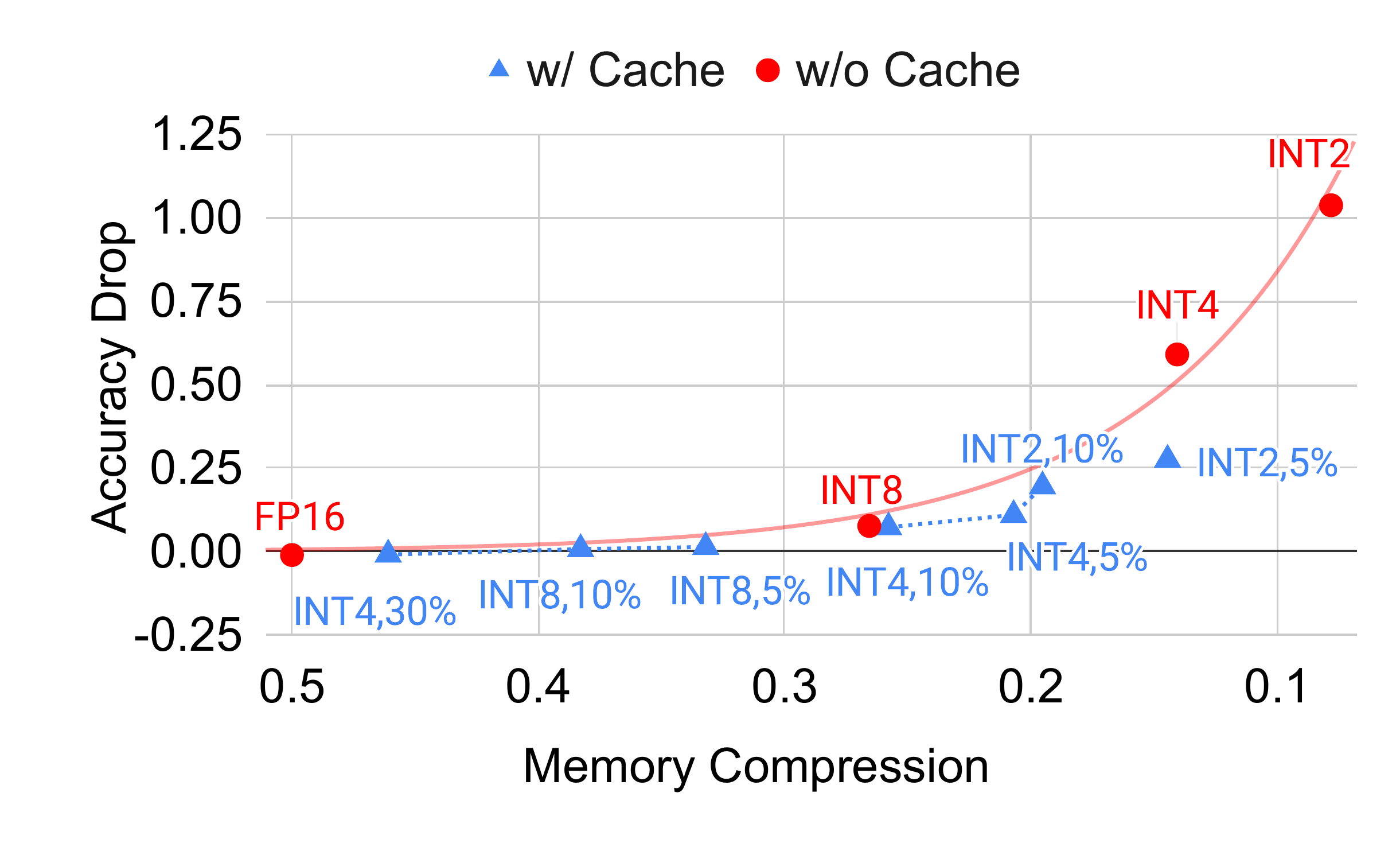}
    \caption{Accuracy vs. memory compression factor with and without high-precision cache on Criteo-Kaggle dataset.}
    \label{fig:cpu-acc-reduction-complete}
\end{figure*}

Figure~\ref{fig:cpu-acc-reduction-complete} plots the complete accuracy vs. memory compression trends for CPU experiments with DLRM-Kaggle dataset using compression factors computed by Table~\ref{appendix:compression}.

\end{document}